\documentclass{article} 
\usepackage{iclr2026_conference,times}

\usepackage{amsmath,amsfonts,bm}

\def\eqref#1{equation~\ref{#1}}

\def\1{\bm{1}}

\DeclareMathAlphabet{\mathsfit}{\encodingdefault}{\sfdefault}{m}{sl}
\SetMathAlphabet{\mathsfit}{bold}{\encodingdefault}{\sfdefault}{bx}{n}

\usepackage[most]{tcolorbox}

\usepackage{hyperref}
\usepackage{url}
\usepackage{wrapfig}
\usepackage{algorithm}
\usepackage{booktabs} 
\usepackage{tabularx} 
\usepackage{multirow}
\usepackage{algpseudocode}
\usepackage{float}
\usepackage{setspace}
\usepackage{threeparttable}
\usepackage[normalem]{ulem}
\usepackage[table]{xcolor}   
\usepackage{arydshln, colortbl}

\usepackage{xcolor}
\usepackage{colortbl}
\usepackage{soul}

\tcbuselibrary{breakable}

\definecolor{SDEblue}{RGB}{28 58 88}
\definecolor{lighterblue}{HTML}{f2fafd}  
\definecolor{cc3}{RGB}{255, 191, 0}
\definecolor{cc4}{RGB}{0, 128, 128}
\hypersetup{
 colorlinks=true,
 linkcolor=SDEblue,   
 urlcolor=magenta,
 citecolor=SDEblue,
}

\newcommand{\sizenormsmall}{\fontsize{9.1pt}{11pt}\selectfont}

\definecolor{customorange}{HTML}{ffe6d5} 
\newcommand{\hlorange}[1]{{\sethlcolor{customorange}\hl{#1}}}

\definecolor{custompink}{HTML}{ffd5d5} 
\newcommand{\hlpink}[1]{{\sethlcolor{custompink}\hl{#1}}}

\newcommand{\hlgray}[1]{{\sethlcolor{gray!15}\hl{#1}}}

\newcommand{\hlgreen}[1]{{\sethlcolor{green!15}\hl{#1}}}

\newcommand{\hlblue}[1]{{\sethlcolor{blue!15}\hl{#1}}}

\title{StreamingThinker: Large Language Models Can Think While Reading}

\author{%
  Junlong Tong\textsuperscript{1,2,3} \quad
  Yingqi Fan\textsuperscript{2} \quad
  Anhao Zhao\textsuperscript{2,4} \quad
  Yunpu Ma\textsuperscript{5} \quad
  Xiaoyu Shen\textsuperscript{2,3}\thanks{Corresponding author} \quad\\
  \textsuperscript{1}Shanghai Jiao Tong University \quad
  \textsuperscript{2}Eastern Institute of Technology, Ningbo\quad\\
  \textsuperscript{3}Ningbo Key Laboratory of Spatial Intelligence and Digital Derivative, Institute of Digital Twin\\
  \textsuperscript{4}Hong Kong Polytechnic University\quad
  \textsuperscript{5}Munich Center for Machine Learning, LMU\\
  \texttt{jl-tong@sjtu.edu.cn}\quad \texttt{xyshen@eitech.edu.cn}
  \vspace{-0.5em}
}

\iclrfinalcopy 
\begin{document}

\maketitle

\begin{abstract}
Large language models (LLMs) have demonstrated remarkable capabilities in chain of thought (CoT) reasoning. However, the current LLM reasoning paradigm initiates thinking only after the entire input is available, which introduces unnecessary latency and weakens attention to earlier information in dynamic scenarios. Inspired by human cognition of thinking while reading, we first design a \textit{\textbf{streaming thinking}} paradigm for LLMs, where reasoning unfolds in the order of input and further adjusts its depth once reading is complete.
We instantiate this paradigm with \textit{StreamingThinker}, a framework that enables LLMs to think while reading through the integration of streaming CoT generation, streaming-constraint training, and streaming parallel inference.
Specifically, StreamingThinker employs streaming reasoning units with quality control for CoT generation, enforces order-preserving reasoning through streaming attention masks and position encoding, and leverages parallel KV caches that decouple input encoding from reasoning generation, thereby ensuring alignment and enabling true concurrency. We evaluate StreamingThinker on the Qwen3 model family across math reasoning, logical reasoning, and context-based QA reasoning tasks.
Experimental results show that the StreamingThinker preserves performance comparable to batch thinking, while yielding an 80\% reduction in token waiting before the onset of reasoning and a more than 60\% reduction in time-level latency for producing the final answer, demonstrating the effectiveness of the streaming paradigm for LLM reasoning.
Code is publicly available at \href{https://github.com/EIT-NLP/StreamingLLM/tree/main/StreamingThinker}{\color{SDEblue}{{this repository.}}}

\end{abstract}

\vspace{-1em}
\section{Introduction}

Large language models (LLMs) have shown impressive reasoning capabilities, as exemplified by systems like OpenAI-o1~\citep{jaech2024openai} and DeepSeek-R1~\citep{guo2025deepseek}. Yet, most current approaches follow a \textit{\textbf{batch thinking}} paradigm, in which reasoning begins only after the entire input context has been received. This paradigm is problematic in scenarios that demand timely responses or dynamic information processing. First, waiting for the full input before reasoning introduces unnecessary latency. Second, as the input increases, attention to earlier information becomes diluted due to the growing disconnect between reasoning steps and their relevant context~\citep{liu2024lost,levy2024same,zhang2025attention}. This weakens coherence and raises the risk of hallucination. To compensate, LLMs often rely on longer chains of thought (CoT)~\citep{wei2022chain,yeo2025demystifying,chen2025towards} or repeated self-refinement~\citep{wangself,ling2023deductive,madaan2023self} to re-focus, which in turn raise computational costs and inflate token usage.

\begin{figure}[t]
\begin{center}
\vspace{-0.8em}
\includegraphics[width=0.98\linewidth]{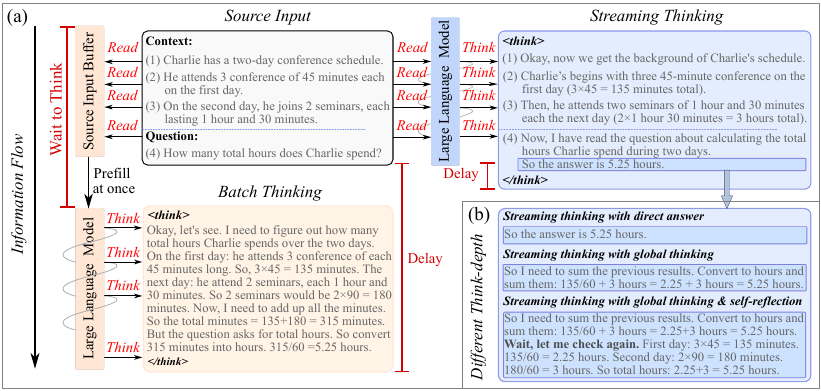}
\end{center}
\vspace{-1.5em}
\caption{\small(a) Standard LLM reasoning follows the \textit{\textbf{batch thinking}} paradigm, where reasoning begins only after the entire input is received, leading to high latency and imbalanced attention to the input. The proposed \textit{\textbf{streaming thinking}} paradigm enables LLMs to think while reading during input reception, substantially reducing latency and maintaining attention aligned with the order of input. (b) Streaming thinking paradigm supports multi-depth reasoning, balancing latency with performance.}

\vspace{-1.2em}

\label{Fig:Main}
\end{figure}

In contrast, human reasoning often unfolds in an immediate and streaming manner. Research in psychology and cognitive science shows that during reading, humans process incoming information instantaneously, including text decoding, meaning construction, background knowledge activation, integrative reasoning, and actively generating inferences to build a coherent understanding~\citep{kintsch1988role, graesser1994constructing}. This \emph{“thinking while reading”} mechanism not only enhances processing efficiency, but also allows reasoning to occur closely alongside the relevant context, minimizing cognitive lag and mitigating the risk of coherence degradation.

To narrow the gap between LLM and human reasoning, we propose a \textit{\textbf{streaming thinking}} paradigm for LLMs. Streaming thinking unfolds reasoning steps alongside the input stream, allowing the model to reason while receiving information. Once the full input is available, the model can further refine its reasoning and adjust the depth of its analysis to match task complexity.\footnote{Appendix~\ref{Appendeix:Applications} discusses the value and potential applications of human-like streaming thinking in practice.} As illustrated in Figure~\ref{Fig:Main}, compared with batch thinking, streaming thinking enables much faster responses while preserving consistency with the order of incoming information.

Accordingly, we propose \textit{StreamingThinker}, a framework that instantiates the streaming thinking paradigm.
The StreamingThinker integrates a streaming CoT generation pipeline with training and inference frameworks that adapt LLMs to the streaming paradigm.
The generation pipeline inserts boundary tokens to the inputs to define minimal reasoning units, prompting LLMs to generate serialized reasoning segments for each unit that are reconstructed into incremental content, filtered by quality evaluation, with reasoning depth controlled through token intervention~\citep{wu2025effectively}.
To support streaming training, StreamingThinker introduces two modifications: a streaming attention mask that restricts each reasoning step to past and current input, and a streaming position encoding that independently indexes input and reasoning tokens from zero to eliminate positional contention, ensuring alignment with associated inputs.
For inference, StreamingThinker employs parallel KV caches that decouple input encoding from reasoning generation and merge only during cross-attention, enabling true thinking while reading.
We conduct a comprehensive evaluation on Qwen3 model family~\citep{yang2025qwen3}, covering diverse tasks such as math reasoning, logical reasoning, and context-based QA reasoning.
Experimental results indicate that StreamingThinker achieves reasoning performance on par with batch thinking, yet reduces token-level waiting before reasoning by 80\% and overall answer latency by over 60\%.

The contributions of this work are fourfold.

\vspace{-0.8em}
\begin{itemize}
    \item
    To the best of our knowledge, we are the first to introduce the streaming thinking paradigm for large language model reasoning. This paradigm mirrors human cognitive processes, enabling LLMs to engage in more timely and continuous thinking in dynamic scenarios.

    \item
    We propose a streaming CoT generation pipeline for this paradigm. Drawing on the principles of human streaming thinking, it ensures that the reasoning process remains aligned with the sequential order of the input context.

    \item

    We provide an adaptation training and inference framework that implements the streaming thinking paradigm, in which training ensures alignment with sequential inputs and inference enables efficient concurrent reasoning.
    \item
    Extensive experiments on diverse reasoning tasks show that our method achieves reasoning performance comparable to batch thinking, while markedly reducing waiting latency.
\end{itemize}

\newpage
\section{Streaming Thinking Paradigm}

\paragraph{Paradigm Design}

Human streaming cognition involves two complementary processes: rapidly generating and updating representations as input arrives, and subsequently performing global integration to transform local, shallow understanding into holistic, deep comprehension~\citep{kintsch1988role}.

Inspired by this process, we design a streaming thinking paradigm for LLMs, with an example illustrated in Figure~\ref{Fig:Main}. At each step, the model incrementally processes the incoming sentence, focusing on progressive comprehension such as (1) understanding and summarizing key information, (2) explaining ambiguities and reorganizing semantic relations, (3) extending logical implications, and (4) skipping thinking step when the content is irrelevant to the question. After completing this incrementally reading and reasoning, we define multiple reasoning depths as a post-reasoning step, as shown in Figure~\ref{Fig:Main} (b). The model may (1) directly produce an answer, representing the shallowest depth; (2) further integrate global information to achieve deeper comprehension; or (3) perform reflective reasoning on top of global integration to obtain the most reliable reasoning outcome.
Within the paradigm, reasoning depth adapts to question complexity and is explicitly controlled by instruction signals that guide the model toward different strategies.

\paragraph{Ordering of Context and Question in Streaming Thinking}

In the batch thinking paradigm, all input is available simultaneously, so the order of context and question is often overlooked. Yet prior work shows that their relative order can influence reasoning~\citep{chen2024premise,wei2024unveiling,xie2024order}, an effect amplified in streaming scenarios. In human reading, two natural input orders commonly occur. In the first, the question is presented before the context, enabling the reader to establish targeted associations as subsequent information is processed. In the second, the context precedes the question, in which case the question remains unavailable during streaming reasoning and the reader must rely solely on the context itself to construct plausible inferences.
To approximate these scenarios, our streaming thinking paradigm explicitly distinguishes between the two orders and examines their impact on model reasoning. Some examples are provided in Appendix~\ref{Appendeix:Paradigm}.

\paragraph{Formal Definition}

Streaming thinking is defined as an immediate reasoning process that unfolds alongside the input stream, with reasoning depth flexibly adapting to the complexity of the problem.
Formally, let $Q$ denote the input question and $C_t$ the $t$-th sentence in the input context.
At each step, the LLM generates an intermediate reasoning state $R_t$ corresponding to $C_t$,
and $R_q$ corresponding to the question $Q$.
The instruction $I$ sets the reasoning depth, directing how intermediate states are integrated into the final reasoning output $R$.
The streaming thinking process can be described as:

{\small
\vspace{-1em}
\begin{align}
    \label{eq:paradigm}
    \mathcal{P}_{\mathrm{streaming}}=
    \begin{cases}
        {\color{SDEblue}{\textstyle\prod_{t=1}^{T}P(R_t|C_{\leq t},R_{\leq t-1})\cdot P(R_q|Q,C_{\leq T},R_{\leq T})}}\cdot P(R|Q,C_{\leq T},R_{\leq T},I), &\text{context~first}, \\
        \\
        \underbrace{\color{SDEblue}{P(R_q|Q)\cdot\textstyle\prod_{t=1}^{T}P(R_t|Q,C_{\leq t},R_{\leq t-1})}}_{\text{\sizenormsmall streaming thinking}}\cdot \underbrace{P(R|Q,C_{\leq T},R_{\leq T},I)}_{\text{\sizenormsmall with controllable depth}}, &\text{question~first}.
    \end{cases}
\end{align}
\vspace{-2em}
}

\section{StreamingThinker}
This section introduces the StreamingThinker, a supervised fine-tuning framework that integrates streaming CoT generation with streaming training and inference mechanisms to adapt batch-oriented LLMs to the streaming thinking paradigm.
\subsection{Streaming CoT Generation}
StreamingThinker first constructs a streaming-like CoT dataset, as existing batch-style reasoning traces lack human-like incremental thinking. This step produces streaming-compatible traces with controllable depth, providing the foundation for subsequent training and evaluation.

\paragraph{Generation Process}

\begin{wrapfigure}{r}{0.43\textwidth}
    \centering

    \includegraphics[width=\linewidth]{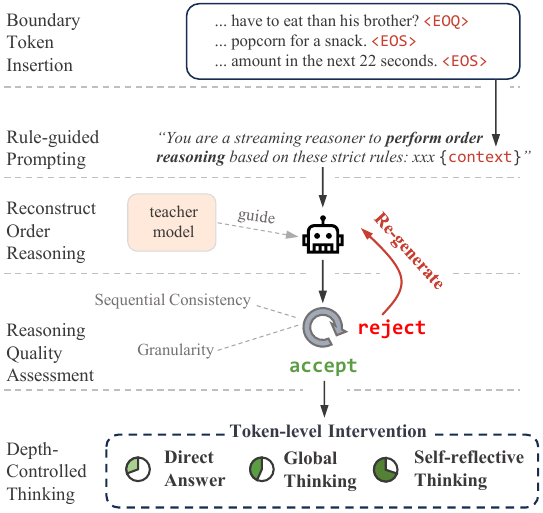}
    \vspace{-1.5em}
    \caption{\small Generation for streaming CoT.}
    \vspace{-2.5em}
    \label{Fig:Data}
\end{wrapfigure}

The streaming reasoning dataset is constructed through a multi-stage pipeline, as shown in Figure~\ref{Fig:Data}. We first insert sentence-level boundary tokens \texttt{<EOS>} for the input to define minimal reasoning units. Then the LLM is prompted to generate order-preserving reasoning for the preceding sentence and terminate the step with \texttt{<EOT>}, when encountering \texttt{<EOS>}.
\footnote{Boundary tokens mark minimal reasoning units and indicate the end of a reasoning step during inference.} To further enforce sequential alignment, a higher-parameter teacher model reconstructs the generated reasoning. Once all sentence-level reasoning traces are generated, they are evaluated using granularity score and sequential consistency score. Passing samples are enhanced with token-level intervention to generate depth-controlled reasoning variants. Samples that fail the evaluation are regenerated, and those still failing under Pass@2~\citep{chen2021evaluating} metric are discarded.

\paragraph{Quality Assurance and Evaluation}

We propose two evaluation metrics: the \textit{granularity score} measures the fine-grained alignment of the streaming CoT, while the \textit{sequential consistency score} assesses whether the reasoning proceeds in a streaming, order-preserving way.
The granularity score is defined as the ratio between the number of boundary tokens in the input and those in the output: $granularity = \frac{N_{\text{EOS}}}{N_{\text{EOT}}}$, where $N_{\text{EOS}}$ and $N_{\text{EOT}}$ denote the counts of boundary tokens in the input and output, respectively.
When the granularity score is equal to 1, the reasoning matches the input in boundary count, suggesting ideal alignment.
The consistency score is defined as the similarity between the input sentence and reasoning sentences, i.e., $consistency = \text{sim}(R_t,C_t) = \frac{v_R \cdot v_C}{\|v_R\| \, \|v_C\|}$, where $v_R$ and $v_C$ denote the embedding vectors of the reasoning sentences $R_t$ and the input sentence $C_t$, respectively. We use SentenceBERT~\citep{reimers2019sentence} for sequential consistency calculation.\footnote{In cases where a single input sentence corresponds to multiple reasoning sentences, we regard them collectively as one segment. We provide the similarity map between the input and the reasoning in the Appendix~\ref{Appendix_Similarity}.}

\subsection{Streaming Training Framework}

A naive approach is to interleave input and reasoning sentences. While it appears to be streaming, this method is inconsistent with the pretraining format of LLMs and still enforces serial execution, where reasoning prevents the model from simultaneously consuming new inputs. Beyond interleaving, we design an actual streaming training framework for streaming thinking paradigm.

\paragraph{Streaming Attention Mask Matrix}
According to Equation~\ref{eq:paradigm}, the core constraint of streaming thinking is that the reasoning step at current time must not access future inputs. In contrast, the standard attention mask used for batch thinking exposes all inputs to every reasoning step. To adapt LLMs to the streaming paradigm, we inject a streaming constraint into the attention mask. As illustrated in Figure~\ref{Fig:Method} (a), within the attention from the reasoning sentences to the input sentences, we apply a causal mask that blocks attention from step $t$ to input positions $>t$. We refer to the masked region as the streaming mask region.
Let the input sentence have length $T$ and the reasoning segment length $L$. The streaming mask is then defined as
{\small
\begin{align}
    \label{eq:mask}
    \mathcal{M}_{\text{streaming}}(i,j)
= \mathcal{M}(i,j) + \bigl(-\infty - \mathcal{M}(i,j)\bigr)\cdot
\mathbb{I}_{\{\,i > T,\ j < T,\ j > i - T + 1\,\}},
\end{align}
}where $\mathcal{M}$ is the vanilla casual mask matrix of LLMs, and $\mathbb{I}$ is an indicator function.

\paragraph{Streaming Position Encoding}

The RoPE~\citep{su2024roformer} of LLMs represents relative positions by rotating queries and keys, with the attention between a reasoning token $R_t$ and an input token $S_t$ expressed as $Attn(R_t, S_t)= q_R^TR(T+t-t)k_S$, where $T$ is the input length and $t,t+T$ are their positional IDs, and $R(T)$ is the rotary matrix.
However, in streaming scenarios, the concurrent generation of output and reception of input induces positional contention in the encoding space~\citep{guo2024decoder,tong2025llm}.
To address this issue, we assign independent position IDs to input and reasoning tokens, both starting from zero.
Formally, the positional IDs of reasoning token $R_t$ and input token $S_t$ are both set to $t$, yielding $Attn(R_t,S_t)= q_R^TR(t-t)k_S$.
This design removes positional contention in streaming parallel processing. Furthermore, identical position IDs ensure that, during streaming reasoning, a reasoning sentence is positioned nearest to its associated input sentence and distant from others, which conforms to the essential principle of streaming alignment.

\subsection{Streaming Inference}
\begin{figure}[t]
\begin{center}
\includegraphics[width=\linewidth]{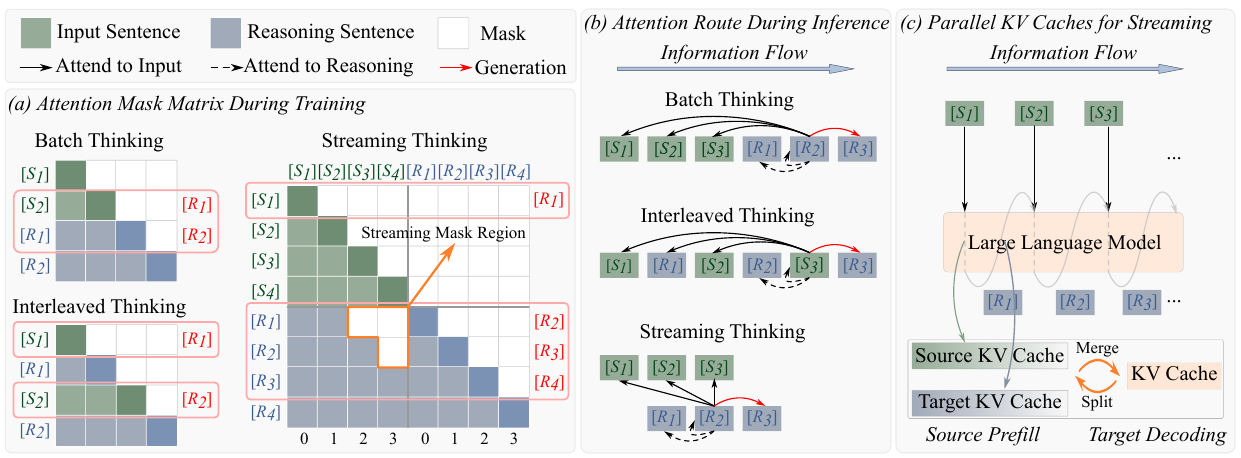}
\end{center}
\vspace{-0.8em}
\caption{\small Training and inference framework of StreamingThinker. (a) shows attention mask at training. (b) and (c) show attention routing and parallel KV caches for streaming thinking at inference.}
\vspace{-0.5em}
\label{Fig:Method}
\end{figure}

\paragraph{Attention Route}
Figure~\ref{Fig:Method}(b) compares information flow across paradigms. In batch thinking, reasoning begins only after the full context is received, resulting in long attention routes and serial dependency. Interleaved thinking alternates reasoning with partial inputs but still updates a single cache sequentially, and its format diverges from the pretraining distribution. In contrast, streaming attention preserves consistency with batch-style pretraining while employing parallel caches, enabling concurrent processing with shorter routes and lower latency.

\paragraph{Parallel KV caches}
To enable parallel processing in streaming reasoning, we design two KV caches during inference: a source cache for input tokens and a target cache for reasoning tokens, as shown in Figure~\ref{Fig:Method} (c). As input arrives sentence by sentence, the LLM performs prefill in arrival order, storing hidden states in the source cache. Before decoding, the two caches are merged so that reasoning can attend to the inputs, and newly generated tokens are written into the merged cache. After finishing a sentence, the caches are split again. This design enables concurrency between source-side prefill and target-side decoding, whereas batch and interleaved paradigms rely on a single continuous cache, enforcing strictly serial execution.

\section{Experiments}
\label{Sec_Experiments}
\subsection{Experimental Settings}
\paragraph{Datasets}

To evaluate StreamingThinker, we conduct a comprehensive assessment across three representative reasoning tasks: math reasoning, logical reasoning, and context-based QA reasoning. For math reasoning, we selected GSM-symbolic~\citep{mirzadehgsm} and MetamathQA~\citep{yumetamath}. For logical reasoning, we utilized LogicNLI~\citep{tian2021diagnosing} and ProofWriter~\citep{tafjord2020proofwriter}. Finally, for context-based QA reasoning, PubMedQA~\citep{jin2019pubmedqa} and HotpotQA~\citep{yang2018hotpotqa} were employed. All datasets were partitioned into dedicated training and testing sets, with detailed specifications provided in the Appendix~\ref{Appendix_Dataset}.

\paragraph{Models and Baselines}
We implement our StreamingThinker using models from the Qwen3 family. For streaming CoT generation, we utilize Qwen3-32B as the initial generation model to produce the preliminary streaming reasoning trace. We assign Qwen3-235B-A22B-Instruct as the teacher guidance model to reconstruct the preliminary trace. Then Qwen3-1.7B and Qwen3-4B as the backbone of StreamingThinker for evaluation.
To provide a comprehensive evaluation, we compare StreamingThinker with three baselines representing alternative reasoning paradigms:
(1) \textit{batch thinking (Batch, orignal)}, where the model generates reasoning after observing the entire context without additional supervision;
(2) \textit{batch thinking with CoT distillation (Batch, SFT)}, where reasoning traces are distilled from a stronger 32B teacher model to enhance reasoning ability; and
(3) \textit{interleaved mode}, a naive streaming variant that alternates between input segments and reasoning steps without parallel cache support.

\paragraph{Metric}
The evaluation of streaming scenarios can be viewed as a trade-off between performance and latency. For reasoning performance, we adopt Pass@1 score as the accuracy metric to measure the model’s ability to successfully solve problems. Latency is assessed at two levels: token latency and time latency. At the token level, we use token-to-first-token (TTFT) to measure how many input tokens must be observed before the model begins reasoning.\footnote{This is similar with time-to-first-token, but measured in terms of token count.}
For time latency, \textcolor{black}{we set the LLM's input speed as the average human speaking rate about 150 words per minute}~\citep{geva2006reading,jacewicz2010between}, and define the waiting time until the first answer token as the latency.\footnote{Detailed definitions and additional evaluation metrics are provided in Appendix~\ref{Appdenx_Metric}.}

\subsection{Effectiveness of the Streaming Thinking Paradigm for LLMs}

We begin by validating the feasibility of the streaming thinking paradigm (sequential reasoning with depth adjustment) under the batch setting, which removes interference from streaming input and cache strategies. This controlled setup allows us to assess the model’s adherence to the paradigm and to further investigate two key aspects: (1) the impact on reasoning trajectories and reasoning depth, and (2) the role of streaming position encoding in maintaining alignment and stability.
\begin{table}[t]
    \centering
    \scriptsize
    \setlength{\tabcolsep}{3.8pt}
    \renewcommand{\arraystretch}{1.3}
    \vspace{-1.5em}
    \caption{ Pass@1 accuracy (Acc↑) and token usage (Tokens↓) results of the streaming thinking paradigm under the batch processing setting. The comparison includes: (1) the original batch thinking baseline without distillation, (2) SFT models trained with RoPE or Streaming RoPE (SPE) distilled from Qwen3-32B CoT data, and (3) the streaming thinking model executed in a batch processing mode (Batch-S) with RoPE or SPE. Reasoning depth is categorized into three levels, denoted as D1–D3, where D1 = direct answer, D2 = with global reasoning, and D3 = with self-reflection.}
    \label{tab:main_results}
    \begin{threeparttable}
    \begin{tabular}{lcccccccccccc}
        \toprule
         & \multicolumn{4}{c}{\textbf{Math reasoning}} & \multicolumn{4}{c}{\textbf{Logical reasoning}} & \multicolumn{4}{c}{\textbf{Context-based QA reasoning}} \\
         \cmidrule(lr){2-5} \cmidrule(lr){6-9} \cmidrule(lr){10-13}
        \textbf{Methods} & \multicolumn{2}{c}{\textbf{GSM-Symbolic}} & \multicolumn{2}{c}{\textbf{MetaMathQA}} & \multicolumn{2}{c}{\textbf{ProofWriter}} & \multicolumn{2}{c}{\textbf{LogicNLI}} & \multicolumn{2}{c}{\textbf{HotPotQA}} &\multicolumn{2}{c}{\textbf{PubMedQA}} \\
        \cmidrule(lr){2-3} \cmidrule(lr){4-5} \cmidrule(lr){6-7} \cmidrule(lr){8-9} \cmidrule(lr){10-11} \cmidrule(lr){12-13}
        & Acc↑ & Tokens↓ & Acc↑ & Tokens↓ & Acc↑ & Tokens↓ & Acc↑ & Tokens↓ & Acc↑ & Tokens↓ & Acc↑ & Tokens↓ \\
        \midrule
        \rowcolor{gray!15} \multicolumn{13}{c}{\textit{Qwen3-1.7B}} \\
        Batch, Original    & 0.645 & 1714.25 & 0.657 & 1751.40 & 0.488 & 1379.58 & 0.503 & 1384.47 & 0.484 & 638.91 & 0.616 & 619.92 \\
        \hdashline
        Batch, SFT, RoPE   & \textbf{0.748} & 1104.05 & 0.755 & 1236.76 & 0.785 & 980.12  & 0.600 & 1120.38 & 0.530 & 512.42 & 0.642 & \textbf{602.38} \\
        Batch, SFT, SPE    & 0.740 & \uline{1032.71} & \textbf{0.761} & \uline{1195.68} & \uline{0.790} & \uline{970.44}  & \textbf{0.604} & \textbf{1110.27} & \uline{0.528} & \textbf{498.73} & \uline{0.648} & 611.25 \\
        \hdashline
        \addlinespace[0.4ex]
        Batch-S, D1, RoPE  & 0.396 & 217.33  & 0.1795 & 253.26  & 0.470 & 360.18  & 0.515 & 872.33  & 0.456 & 324.71 & 0.598 & 452.15 \\
        Batch-S, D1, SPE   & 0.401 & 212.42  & 0.183  & 250.26  & 0.476 & 352.07  & 0.519 & 865.27  & 0.463 & 318.64 & 0.601 & 448.92 \\
        \hdashline
        Batch-S, D2, RoPE  & 0.688 & 396.34  & 0.703  & 611.61  & 0.540 & 555.21  & 0.542 & 1196.33 & 0.492 & 472.51 & 0.624 & 571.66 \\
        Batch-S, D2, SPE   & 0.692 & 391.25  & 0.706  & 618.53  & 0.544 & 545.37  & 0.546 & 1184.47 & 0.497 & 465.33 & 0.628 & 565.92 \\
        \hdashline
        Batch-S, D3, RoPE  & \uline{0.732} & 562.88  & 0.736  & \textbf{829.58}  & 0.804 & 720.46  & 0.560 & \uline{1351.16} & 0.545 & 589.42 & \textbf{0.668} & 693.18 \\
        Batch-S, D3, SPE   & 0.727 & \textbf{552.13}  & \uline{0.738}  & 835.11  & \textbf{0.810} & \textbf{710.32}  & \uline{0.568} & 1362.82 & \textbf{0.552} & \uline{583.61} & 0.663 & \uline{688.74} \\
        \midrule
        \rowcolor{gray!15} \multicolumn{13}{c}{\textit{Qwen3-4B}} \\
        Batch, Original    & 0.855 & 1445.61 & 0.774 & 1630.67 & 0.620 & 1878.57 & 0.492 & 1421.92 & 0.575 & 617.40 & 0.609 & 463.37 \\
        \hdashline
        Batch, SFT, RoPE   & \textbf{0.890} & 896.45 & 0.833 & 1029.66 & \textbf{0.863} & 935.36 & 0.647 & 1002.37 & \uline{0.608} & \textbf{455.56} & 0.675 & \uline{572.63} \\
        Batch, SFT, SPE    & 0.882 & \uline{887.14} & \textbf{0.836} & \uline{989.23} & 0.861 & \uline{925.66} & \uline{0.650} & \textbf{992.38} & 0.601 & 467.21  & \textbf{0.680} & 582.31 \\
        \hdashline
        Batch-S, D1, RoPE  & 0.433 & 203.23 & 0.662 & 201.60 & 0.592 & 386.25 & 0.627 & 683.17 & 0.552 & 269.45 & 0.598 & 358.14 \\
        Batch-S, D1, SPE   & 0.437 & 199.64 & 0.668 & 200.73 & 0.596 & 376.13 & 0.623 & 672.61 & 0.561 & 257.11 & 0.592 & 351.47 \\
        \hdashline
        Batch-S, D2, RoPE  & 0.864 & 348.52 & 0.791 & 534.74 & 0.801 & 583.28 & 0.639 & 983.52 & 0.592 & 381.36 & 0.657 & 507.32 \\
        Batch-S, D2, SPE   & 0.871 & 352.17 & 0.806 & 528.33 & 0.795 & 596.61 & 0.642 & 978.52 & 0.601 & 388.93 & 0.651 & 498.17 \\
        \hdashline
        Batch-S, D3, RoPE  & 0.867 & 499.83 & 0.821 & \textbf{661.96} & 0.856 & 721.89 & 0.645 & 1242.82 & 0.616 & 474.56 & 0.661 & \textbf{563.41} \\
        Batch-S, D3, SPE   & \uline{0.874} & \textbf{493.26} & \uline{0.825} & 668.19 & \uline{0.861} & \textbf{715.26} & \textbf{0.651} & \uline{1176.65} & \textbf{0.621} & \uline{466.14} & \uline{0.669} & 571.22 \\
        \bottomrule
    \end{tabular}
    \end{threeparttable}
    \vspace{-1em}
\end{table}

\vspace{-1em}
\paragraph{Effect of Reasoning Depth in Streaming Thinking}

\begin{figure}[t]
\begin{center}
\includegraphics[width=\linewidth]{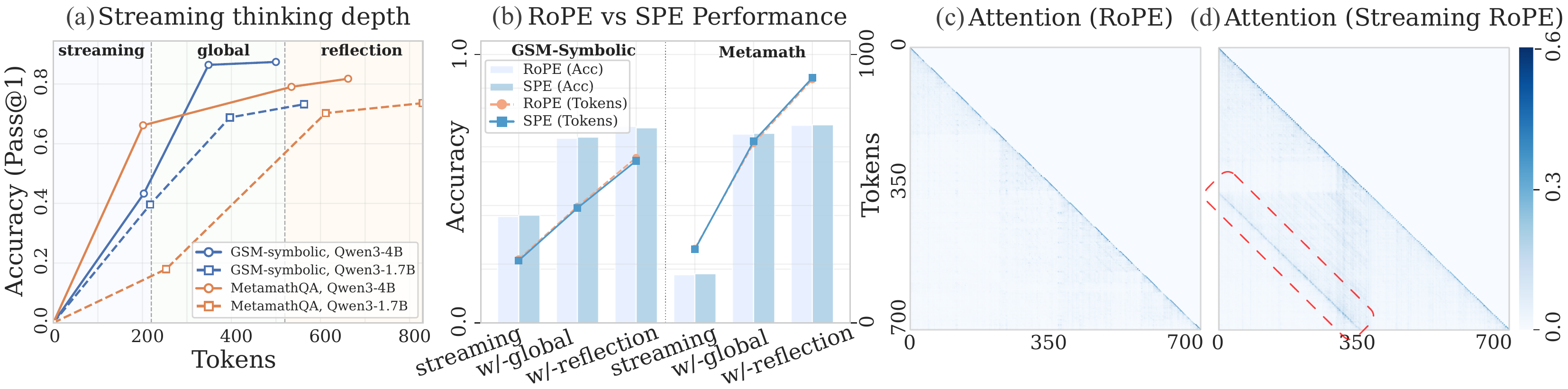}
\end{center}
\vspace{-1.5em}

\caption{{Streaming thinking performance and attention patterns. Subplots (a)–(b) show accuracy–token trade-offs for GSM-Symbolic and MetaMathQA under RoPE and streaming RoPE (SPE); subplots (c)–(d) show average attention maps comparing RoPE and streaming RoPE.}}
\vspace{-1em}
\label{Fig:EX1}

\end{figure}

We first validate the streaming thinking framework on Qwen3-1.7B and Qwen3-4B.
As shown in Table~\ref{tab:main_results} and Figure~\ref{Fig:EX1}, the performance of LLMs improves consistently with increasing reasoning depth in the streaming paradigm. At shallow depths, the model mainly performs local reasoning aligned with the sequential input, which provides fast but relatively coarse-grained understanding. When deeper reasoning stages are introduced—particularly with global reflection—the performance approaches that of batch thinking, demonstrating that additional depth helps compensate for the information fragmentation inherent in streaming reasoning. Moreover, results in Table~\ref{tab:main_results} show that under the batch processing setting, the streaming thinking paradigm achieves performance comparable to the original batch thinking, while demonstrating notable advantages in token efficiency.

The slope of the curves in Figure~\ref{Fig:EX1}(a) highlights the marginal gains of each added depth. Notably, introducing the global reasoning stage leads to the most significant improvement, which aligns with the motivation of streaming thinking: since the streaming phase processes inputs in a lightweight, incremental fashion, a global consolidation step is essential to fully integrate dispersed information and support complex inference.

\paragraph{Position Encoding in Streaming Thinking}
Streaming RoPE assigns consistent yet independent index groups to input and reasoning tokens, ensuring that each reasoning step is correctly aligned with its corresponding input. This design also prevents positional interference, thereby addressing the positional contention issue that arises with the original RoPE in streaming scenarios.

Our experiments further demonstrate that Streaming RoPE achieves performance comparable to the original RoPE under the same settings. As shown in Figure~\ref{Fig:EX1}(b) and Table~\ref{tab:main_results}, in math reasoning tasks at the same depth, Streaming RoPE attains nearly identical accuracy and token consumption to the original RoPE. This indicates that adapting RoPE to the streaming paradigm preserves model capacity without incurring performance degradation.

The attention visualizations in Figures~\ref{Fig:EX1}(c) and (d) provide additional insights. While the original RoPE exhibits no clear positional preference across the input context, Streaming RoPE shows a pronounced diagonal concentration, reflecting stronger focus on the current context. This bias aligns well with the motivation of streaming thinking: reasoning should primarily rely on information already observed, thereby enabling the model to “\textit{think while reading}.”

\subsection{LLMs Thinking While Reading Under the Streaming Thinking Paradigm}
After confirming the feasibility of streaming thinking in batch settings, we now extend our evaluation to real streaming scenarios, where inputs arrive incrementally over time. In this setting, the model is required to perform reasoning online, relying solely on the partial context available at each step.
\begin{table}[t]
\begin{center} \scriptsize
\caption{\small Results compare the original batch thinking (i.e., without distillation process) and streaming thinking paradigms. The streaming thinking results include both the naive interleaved streaming mode (Interleaved) and our proposed parallel streaming mode (Streaming), evaluated under three reasoning depths (D1 = direct answer, D2 = global thinking, D3 = self-reflection). Results are reported in terms of Pass@1 (Acc), token number to first input token (TTFT), and time delay for generating first answer token (delay). All experiments are conducted on Qwen3-4B.}
\setlength{\tabcolsep}{2pt}
\renewcommand{\arraystretch}{1.2}
\begin{tabularx}{\textwidth}{l *{9}{>{\centering\arraybackslash}X}@{}}
    \toprule
    \textbf{Method}  & \multicolumn{3}{c}{\textbf{GSM-Symbolic}} & \multicolumn{3}{c}{\textbf{MetaMathQA}} & \multicolumn{3}{c}{\textbf{ProofWriter}} \\
    \cmidrule(lr){2-4}  \cmidrule(lr){5-7} \cmidrule(lr){8-10}
    & Acc↑ & TTFT↓ & Delay↓ (s)
    & Acc↑ & TTFT↓ & Delay↓ (s)
    & Acc↑ & TTFT↓ & Delay↓ (s)
    \\
    \midrule
    \textcolor{black}{Batch, Original}     & \uline{0.855} & \uline{94.74}  & \uline{47.70} & {0.774} & \uline{100.51}  & \uline{53.81} & \uline{0.620} & \uline{232.11} & \uline{61.99} \\
    \hdashline
    Interleaved, D1   & 0.410 & 20.77  & 6.30  & 0.655 & 16.89   & 6.23  & 0.583 & 20.51  & 11.96 \\
    Interleaved, D2   & 0.829 & 20.77  & 11.90 & 0.754 & 16.89   & 16.55 & 0.750 & 20.51  & 18.07 \\
    Interleaved, D3   & 0.843 & 20.77  & 15.46 & \textbf{0.783} & 16.89   & 20.49 & 0.801 & 20.51  & 22.35 \\
    \hdashline
    Streaming, D1     & 0.421 & 20.77  & \textbf{0.66}  & 0.657 & 16.89   & \textbf{0.68}  & 0.588 & 20.51  & \textbf{0.71} \\
    Streaming, D2     & 0.842 & 20.77  & 5.973 & 0.752 & 16.89   & 10.98 & 0.761 & 20.51  & 6.50 \\
    Streaming, D3     & \textbf{0.856} & \textbf{20.77}  & \textbf{9.768} & \uline{0.780} & \textbf{16.89}   & \textbf{15.18} & \textbf{0.813} & \textbf{20.51}  & \textbf{11.05} \\
    \toprule
    \textbf{Method}  & \multicolumn{3}{c}{\textbf{LogicNLI}} & \multicolumn{3}{c}{\textbf{HotPotQA}} & \multicolumn{3}{c}{\textbf{PubMedQA}} \\
    \cmidrule(lr){2-4}  \cmidrule(lr){5-7} \cmidrule(lr){8-10}
    & Acc↑ & TTFT↓ & Delay↓ (s)
    & Acc↑ & TTFT↓ & Delay↓ (s)
    & Acc↑ & TTFT↓ & Delay↓ (s)
    \\
    \midrule
    \textcolor{black}{Batch, Original}    & \uline{0.492} & \uline{350.08} & \uline{46.92} & \uline{0.575} & \uline{1485.547} & \uline{20.37} & \uline{0.609} & \uline{357.468} & \uline{15.29} \\
    \hdashline
    Interleaved, D1   & 0.591 & 42.06  & 21.17 & 0.537 & 24.32   & 8.33  & 0.571 & 21.74  & 11.09 \\
    Interleaved, D2   & 0.614 & 42.06  & 30.47 & 0.576 & 24.32   & 12.02 & 0.633 & 21.74  & 15.71 \\
    Interleaved, D3   & 0.627 & 42.06  & 38.50 & 0.598 & 24.32   & 14.45 & 0.646 & 21.74  & 17.45 \\
    \hdashline
    Streaming, D1     & 0.603 & 42.06  & \textbf{0.72}  & 0.544 & 24.32   & \textbf{0.69}  & 0.579 & 21.74  & \textbf{0.74} \\
    Streaming, D2     & 0.629 & 42.06  & 9.9   & 0.581 & 24.32   & 3.92  & 0.641 & 21.74  & 4.92 \\
    Streaming, D3     & \textbf{0.634} & \textbf{42.06}  & \textbf{18.45} & \textbf{0.603} & \textbf{24.32}   & \textbf{6.50}  & \textbf{0.653} & \textbf{21.74}  & \textbf{6.76} \\
\bottomrule
\end{tabularx}
\label{Tab:streaming_qc}
\vspace{-2.5em}
\end{center}
\end{table}

\vspace{-0.5em}
\paragraph{Latency of StreamingThinker}
We examine the latency performance of StreamingThinker using the Qwen3-4B model.
As shown in Table~\ref{Tab:streaming_qc}, streaming thinking achieves a markedly lower TTFT than batch reasoning, as reasoning can be initiated once the first input segment becomes available. The minimal latency observed at depth~D1 further indicates that reasoning overlaps with input reading without incurring additional overhead.\footnote{It is important to note that the input rate is set to 150 words per minute to match the average speed of human speech in interactive scenarios. Given that LLM decoding operates at a much faster rate, the effective bottleneck in streaming reasoning stems from input arrival, rather than output generation.}
These results confirm that StreamingThinker substantially reduces response delay, a property of particular importance for streaming applications.

\vspace{-0.5em}
\paragraph{Interleaved mode and Streaming mode}
The interleaved mode constitutes a naive instantiation of streaming reasoning. Relative to batch reasoning, it exhibits lower latency—most prominently in terms of TTFT—as reasoning can be initiated earlier, as reported in Table~\ref{Tab:streaming_qc}. Nevertheless, its accuracy is consistently lower and its overall delay higher than those achieved by streaming thinking. This discrepancy can be attributed to the distributional mismatch between interleaved input sequences and the LLMs’ pre-training corpus, which impairs reasoning fidelity. Moreover, the interleaved paradigm enforces a sequential synchronization constraint, requiring the completion of ongoing reasoning before additional input tokens can be incorporated, thereby exacerbating latency. In contrast, StreamingThinker employs parallelized KV caches that disentangle input encoding from reasoning generation, enabling concurrent reading and reasoning. This architectural design effectively minimizes latency while preserving reasoning quality, thereby highlighting the necessity of streaming-specific mechanisms for efficient online reasoning.

\vspace{-1em}
\subsection{Ordering of Context and Question for StreamingThinker}

\begin{wrapfigure}{r}{0.46\textwidth}
    \centering

    \vspace{-1.5em}
    \includegraphics[width=\linewidth]{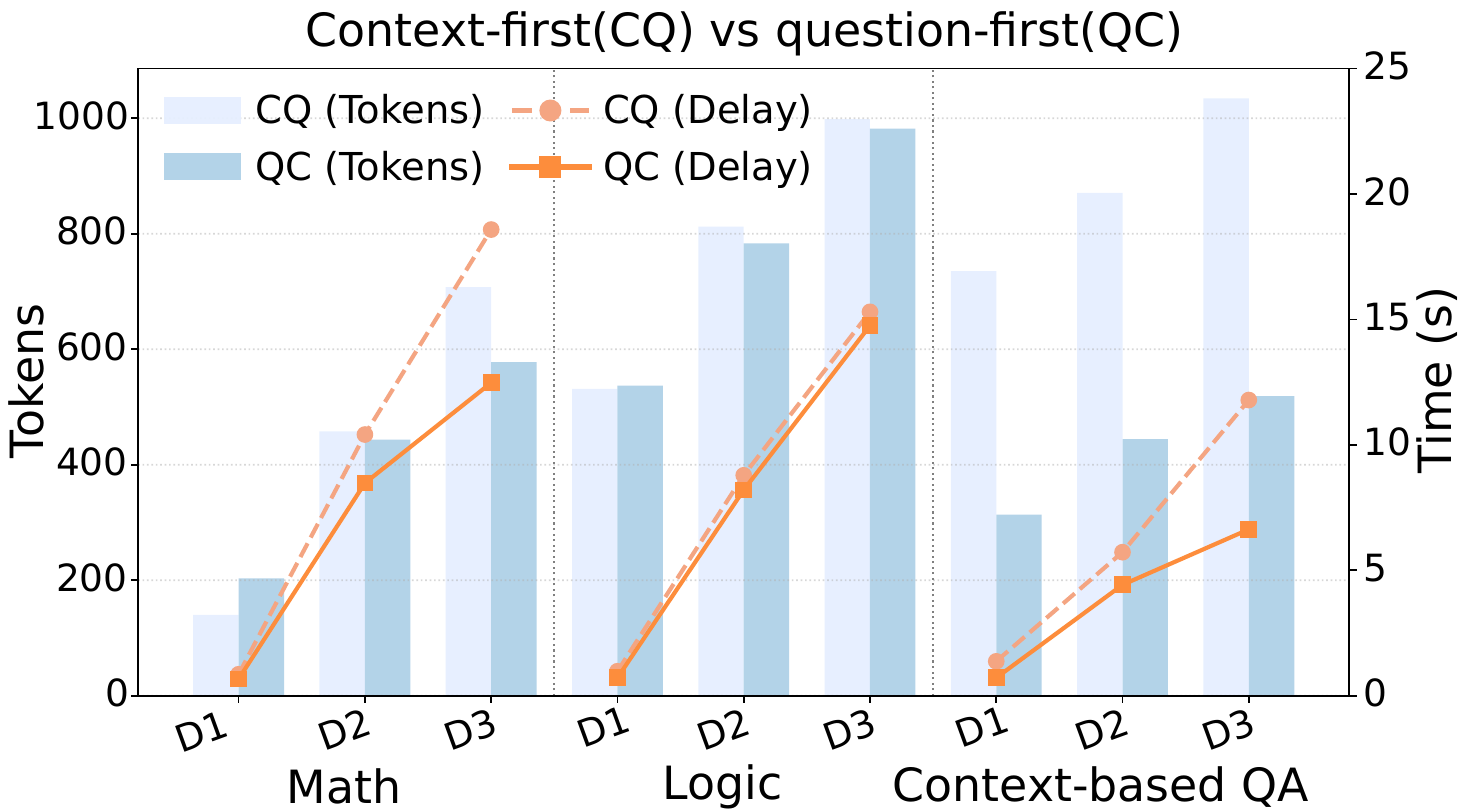}
    \vspace{-2em}

    \caption{\small Comparison between context-first and question-first settings. Bars indicate model token consumption, while lines represent time latency.}

    \vspace{-1.5em}
    \label{Fig_QC_CQ}

\end{wrapfigure}
The ordering of context and question plays a critical role in streaming reasoning. Unlike batch settings, where the model has access to the entire input simultaneously, the streaming paradigm requires reasoning to unfold as inputs arrive. In real-world scenarios, however, the order in which the question and context appear is often unknown. To account for this, we evaluate the proposed streaming thinking framework under both orderings to examine its robustness across different streaming conditions.

Table~\ref{Tab:streaming_cq} reports the performance of StreamingThinker under the context-first input setting across different datasets. Overall, the results follow a similar trend to the question-first setting, confirming the framework’s ability to provide timely responses regardless of input order. As the depth of reasoning increases, both accuracy and latency improve, albeit with a gradual rise in token consumption.

Figure~\ref{Fig_QC_CQ} highlights the differences between the two settings.\footnote{For clarity, we report the average performance for tasks of the same type.} When the question appears first, the model is aware of the reasoning target. This is particularly advantageous in Context-based QA tasks where critical information is sparse; the prior knowledge of the question allows the model to precisely capture key evidence, thereby avoiding the generation of reasoning for irrelevant context (in contrast to domains with denser information). However, for the identified relevant segments, the model tends to expand its logic immediately. This leads to higher token usage at depth D1, and since part of the reasoning is already completed during the streaming process, the incremental token growth at deeper levels (D2 and D3) becomes smaller compared to the context-first setting. In contrast, when the context appears first, the model lacks knowledge of which information is salient. As a result, its reasoning remains more conservative, proceeding sentence by sentence without extensive elaboration, which produces lower token usage at D1 when processing the same context sentence.\footnote{The model’s conservative reasoning stems from our cautious data generation strategy—when the question is unknown, excessive logical expansion may cause overthinking.} However, this characteristic leads to inefficiency in sparse scenarios (such as the context-based QA tasks): due to the conservative strategy, the model fails to skip irrelevant information, resulting in a significantly longer total generation length.

\begin{table}[t]
\begin{center}
\scriptsize
\caption{Results on context-first setting, where the LLM receives context before the question. (D1 = direct answer, D2 = global thinking, D3 = self-reflection). Results are reported in terms of Pass@1 (Acc), token numbers (Token), token number to first input token (TTFT), and time delay for generating first answer token (delay). All experiments are conducted on Qwen3-4B.}
\setlength{\tabcolsep}{2pt}
\renewcommand{\arraystretch}{1.2}
\begin{tabularx}{\textwidth}{l *{20}{>{\centering\arraybackslash}X}@{}}
    \toprule
    \textbf{Method}  & \multicolumn{4}{c}{\textbf{GSM-Symbolic}} & \multicolumn{4}{c}{\textbf{MetaMathQA}} & \multicolumn{4}{c}{\textbf{ProofWriter}} \\
    \cmidrule(lr){2-5} \cmidrule(lr){6-9} \cmidrule(lr){10-13}
    & Acc↑ & Token↓ & TTFT↓ & Delay↓
    & Acc↑ & Token↓ & TTFT↓ & Delay↓
    & Acc↑ & Token↓ & TTFT↓ & Delay↓
    \\
    \midrule
    \textcolor{black}{Batch, Original}     & 0.842 & 1483.65 & \uline{94.74}  & \uline{48.20} & 0.774 & 1668.30 & \uline{100.51} & \uline{54.23} & 0.633 & 1786.42 & \uline{232.11} & \uline{58.05} \\
    \hdashline
    Streaming, D1     & 0.476 & 165.42  & 18.68  & \textbf{0.77}   & 0.668 & 114.50  & 20.40  & \textbf{0.94}  & 0.528 & 457.91  & 8.12   & \textbf{1.06} \\
    Streaming, D2     & 0.837 & 451.81  & 18.68  & 9.23 & 0.760 & 463.59  & 20.40  & 11.59 & 0.740 & 697.20  & 8.12   & 7.33 \\
    Streaming, D3     & 0.844 & 797.63  & \textbf{18.68}  & \textbf{20.45} & 0.788 & 617.36  & \textbf{20.40}  & \textbf{16.71} & 0.798 & 875.19  & \textbf{8.12}   & \textbf{13.11} \\
    \toprule
    \textbf{Method}   & \multicolumn{4}{c}{\textbf{LogicNLI}}     & \multicolumn{4}{c}{\textbf{HotPotQA}}    & \multicolumn{4}{c}{\textbf{PubMedQA}} \\
    \cmidrule(lr){2-5} \cmidrule(lr){6-9} \cmidrule(lr){10-13}
    & Acc↑ & Token↓  & TTFT↓   & Delay↓
    & Acc↑ & Token↓  & TTFT↓   & Delay↓
    & Acc↑ & Token↓  & TTFT↓   & Delay↓
    \\
    \midrule
    \textcolor{black}{Batch, Original}     & 0.483 & 1487.56 & \uline{350.08} & \uline{48.35} & 0.566 & 1492.41 & \uline{1485.55} & \uline{48.50} & 0.593 & {1320.36} & \uline{357.47} &\uline{ 42.90} \\
    \hdashline
    Streaming, D1     & 0.476 & 604.84  & 5.60  & \textbf{0.89} & 0.532 & 1172.36  & 46.39   & \textbf{1.62} & 0.562 & 298.51  & 31.49  & \textbf{1.12} \\
    Streaming, D2     & 0.613 & 927.35  & 5.60  & 10.24 & 0.575 & 1266.41  & 46.39   & 5.90 & 0.628 & 475.63  & 31.49  & 5.54 \\
    Streaming, D3     & 0.625 & 1122.32 & \textbf{5.60}  & \textbf{16.55} & 0.591 & 1381.73  & \textbf{46.39}   & \textbf{12.14} & 0.640 & 687.45  &\textbf{ 31.49}  &\textbf{ 12.43} \\
    \bottomrule
\end{tabularx}
\label{Tab:streaming_cq}
\vspace{-2em}
\end{center}
\end{table}

\section{Discussion}
\paragraph{Efficiency Analysis}
We evaluate the model efficiency on Qwen3-4B using 100 samples from GSM-Symbolic dataset. As shown in Table \ref{tab:efficiency_qwen3}, the Streaming paradigm reduces the first-token latency (measured by time consumption) from 28.00s to 6.23s ($\sim$4.5$\times$ speedup). Crucially, the parallel KV cache operations introduce negligible temporal overhead, with $split_{kv}$ and $merge_{kv}$ taking less than 5ms combined. Additionally, peak memory usage remains consistent with the baseline ($\sim$7.99 GB). While bandwidth cost increases, this is primarily due to the inherent multiple prefill phases in streaming scenarios rather than the parallel KV cache mechanism itself. Note that Table \ref{tab:efficiency_qwen3} reports the latency for each stage separately; the actual end-to-end wall-clock time, which benefits from the concurrent execution of reading and reasoning, has been detailed in Section~\ref{Sec_Experiments}.

\paragraph{Why Streaming Thinker Work?}

Prior studies~\citep{fan2025missing,laban2025llms} have highlighted the risks of reasoning over incomplete inputs.
This phenomenon reflects the inadequacy of batch-processing LLMs when facing local contexts, as their training objectives align with global visibility rather than incremental inference.
However, Streaming Thinker circumvents these pitfalls through three distinct mechanisms. First, global information is \textbf{deferred rather than lost}. Unlike scenarios where critical conditions are permanently removed, our framework merely shifts the timing of acquisition, ensuring the model incorporates the full global context after the streaming phase. Second, we employ a \textbf{conservative reasoning strategy}. Instead of attempting premature complex reflection, the model concentrates on shallow reasoning'' (e.g., intermediate calculations and entity tracking) during the streaming phase. This functions as an incremental pre-processing step that simplifies raw context. Third, this behavior is enforced via a \textbf{specialized streaming training paradigm}. Unlike batch models that rigidly apply full-context patterns to partial inputs—often falling into the over-thinking'' trap—our model is explicitly adapted to local scopes, learning to process available information without jumping to erroneous conclusions.

\begin{table*}[t]

\centering
\caption{Efficiency evaluation on Qwen3-4B. We compare the efficiency of Batch and Streaming paradigms using 100 randomly selected samples from the {GSM-Symbolic} dataset. Metrics include execution memory usage (Peak/$\Delta$), bandwidth cost, and time consumption across different stages. }
\label{tab:efficiency_qwen3}

\renewcommand{\arraystretch}{1.2}
\resizebox{\textwidth}{!}{
\begin{tabular}{l ccc ccccc}
\toprule
\multirow{2}{*}{\textbf{Metric}} & \multicolumn{3}{c}{\textbf{Batch Thinking Paradigm}} & \multicolumn{5}{c}{\textbf{Streaming Thinking Paradigm}} \\
\cmidrule(lr){2-4} \cmidrule(lr){5-9}
 & {Read} & {Prefill} & {Decoding} & {Read} & {Split KV} & {Prefill} & {Merge KV} & {Decoding} \\
\midrule
{Count (times)}  & 69.86 & 1 & 436.0 & 69.86 & 4.65 & 4.65 & 4.65 & 439.0 \\
\hdashline
{Peak Mem (MB)}  & -- & 7,991 & 7,998 & -- & 7,957 & 7,993 & 7,940 & 7,999 \\
{Total Mem $\Delta$ (MB)} & -- & +222 & +102 & -- & -289 & +215 & +217 & +103 \\

{Bandwidth Cost (MB)}   & -- & -- & -- & -- & -- & +29,173 & +434 & -3,310 \\
\hdashline
{Avg Time (s)}   & 0.4 & 0.052 & 0.039 & 0.4 & $<0.001$ & 0.041 & $<0.001$ & 0.037 \\
{Total Time (s)} & 27.945 & 0.052 & 17.178 & 27.945 & \textbf{0.005} & \textbf{0.190} & {0.004} & 16.392 \\
{First-token Latency (s)} & 28.003 & -- & -- & \textbf{6.231} & -- & -- & -- & -- \\
\bottomrule
\end{tabular}
}

\end{table*}

\section{Related Work}
\paragraph{Efficient Reasoning in LLMs}

Prior studies on efficient reasoning in LLMs have mainly focused on three directions: token compression~\citep{aggarwal2025l1,xia2025tokenskip,zhang2025lightthinker,zhao2026policy}, structural quantization and pruning~\citep{liu2025quantization,srivastava2025towards,zhao2025skipgpt}, and efficient decoding~\citep{pan2025learning,liao2025reward}.
Token compression methods such as Tokenskip~\citep{xia2025tokenskip} condense CoT into fewer tokens to reduce the inference cost.
Structural approaches leverage quantization or pruning to compress model parameters for efficient reasoning such as~\citep{srivastava2025towards}.
Efficient decoding has been explored through parallel sampling of generation paths~\citep{pan2025learning} or by using smaller models for speculative prediction~\citep{liao2025reward} to reduce inference latency.
In contrast, our work introduces streaming processing as a complementary dimension of efficiency, allowing reasoning to proceed concurrently with input processing to reduce response latency.
\vspace{-1em}

\paragraph{Streaming LLMs}
Recent research on streaming LLMs has focused on three directions: architecture adaptation~\citep{raffel2024simultaneous,guo2024decoder,tong2025llm}, latency control~\citep{ma2018stacl,ahmed2025non,cheng2025seed}, and modality adaptation~\citep{chen2024videollm,chen2025livecc,xie2024mini,lin2026speak}. Architecture adaptation addresses the mismatch between decoder-only transformers and streaming settings~\citep{tong2025llm}, where issues such as positional interference and redundant re-encoding are mitigated through recurrent states, and modified attention mechanisms. Latency control emphasizes fine-grained alignment between input and output, ranging from fixed policies (e.g., wait-k~\citep{ma2018stacl}) to adaptive scheduling that optimize the accuracy–latency balance. Modality adaptation extends streaming LLMs beyond text to speech recognition~\citep{guo2024decoder}, speech translation~\citep{cheng2025seed}, and video understanding~\citep{chen2024videollm} by integrating modality-specific encoders and synchronization mechanisms. From a reasoning perspective, recent studies~\citep{xie2025interleaved,chiang2025stitch,xie2025mini} have explored alternating reasoning and generation to approximate streaming operation. Our work differs by explicitly modeling thinking while reading, enabling reasoning to evolve concurrently with incremental input.

\section{Conclusion}

\vspace{-1em}
In this work, we introduce the streaming thinking paradigm for large language models, inspired by the human ability to think while reading. Unlike conventional batch reasoning, this paradigm unfolds reasoning concurrently with input arrival and adapts its depth after reading is complete.
To instantiate this paradigm, we developed StreamingThinker, which integrates a streaming CoT generation pipeline, streaming-constrained training, and parallel inference supported by specialized KV cache designs.
Comprehensive experiments across math reasoning, logical reasoning, and long-context QA demonstrate that StreamingThinker substantially reduces latency while preserving or improving reasoning quality. Our analysis further reveals the benefits of controllable reasoning depth, streaming-specific position encoding, and parallel inference for enabling true concurrency. These findings highlight streaming thinking as a promising new direction for efficient and coherent reasoning in LLMs, bridging the gap between artificial and human-like cognition.

\newpage

\bibliography{iclr2026_conference}
\bibliographystyle{iclr2026_conference}

\newpage
\appendix
\setcounter{figure}{0}
\setcounter{table}{0}
\setcounter{equation}{0}

\section{Motivation and Prospective Applications}
\label{Appendeix:Applications}

\vspace{-0.25em}
\subsection{Motivation and Prospective Applications}
One motivation for introducing \textit{streaming thinking} arises from the limitations of the conventional batch reasoning paradigm. In batch reasoning, a model must wait until the entire input is observed before producing any reasoning, which introduces latency, prevents timely responses, and undermines robustness in handling dynamic or sequential information. In contrast, streaming thinking enables large language models to reason incrementally as input arrives, closely mirroring the human cognitive process of \textit{thinking while reading}. A second motivation lies in its broad practical value: this capability unlocks a wide range of applications where timely reasoning and continuous adaptation are essential. We provide some potential applications as follows.

\vspace{-0.25em}
\paragraph{Real-time Dialogue and Interactive Systems}

In advanced conversational agents or AI tutors, streaming thinking allows the model to perform continuous reasoning on a user's partial utterances. For example, it can infer a user's evolving intent, reason about the logical consistency of their arguments in real-time, and formulate clarifying questions or guidance without waiting for the user to pause. This enables a fluid, collaborative dialogue rather than a static, turn-based exchange.
\vspace{-0.25em}
\paragraph{Long-Context Analysis and Synthesis} When processing lengthy documents, live transcripts, or codebases, the model can engage in incremental synthesis. It continuously builds and refines a mental model of the content, reasoning about causal links, logical dependencies, and thematic connections across the information stream. This is crucial for tasks like real-time meeting summarization, where key decisions and action items must be identified and reasoned about as they emerge.
\vspace{-0.25em}
\paragraph{Human-AI Collaborative Reasoning} In creative and analytical workflows, streaming thinking facilitates a true partnership. An AI can act as a thought partner, reasoning alongside a human analyst or writer. As the human proposes an idea, the AI can immediately reason about its implications, offer counter-arguments, or synthesize it with prior information, creating a dynamic and interactive brainstorming loop that accelerates discovery and innovation.
\vspace{-0.25em}
\paragraph{Dynamic Decision-Making and Planning} In high-stakes environments such as autonomous navigation or real-time financial market analysis, streaming thinking is critical. An autonomous agent must reason about a constantly changing environment from a stream of sensory data to make timely decisions. This involves continuously updating its world model, predicting future states, and re-evaluating its action plans based on the most current information, a process impossible under the latency of a batch thinking paradigm.
\vspace{-0.25em}
\paragraph{Embodied Intelligence and Robotic Control} The streaming thinking paradigm is also particularly well-suited for the challenges presented by embodied AI. An embodied agent, like a robot, operates within a dynamic physical environment, which distinguishes it from models that process static information. It continually receives multi-modal sensory data
and is required to respond to this information in a timely manner. In this context, streaming thinking allows the agent to engage in continuous perceptual reasoning, interpreting incoming data to consistently update its internal model of the world. This capability supports dynamic instrumental reasoning, where the model can flexibly plan and re-plan its actions to navigate changing conditions and work towards a goal. For example, a household robot would need to reason about multiple factors at once, such as the movement of a person, the delicacy of an object it plans to handle, and the best path through a cluttered space, while adapting its motor commands accordingly. This close coupling of perception, reasoning, and action in a real-time loop is a core aspect of embodied cognition, and achieving it presents significant challenges for a conventional batch thinking approach.

\vspace{-0.25em}
\paragraph{Streaming Multimodal Understanding}
Beyond text, streaming thinking is vital for interpreting continuous non-verbal data streams, such as live video feeds or audio environments. For instance, in video understanding, a model must reason about the temporal causality of events as they occur—identifying that an action in frame $t$ is a consequence of an event in frame $t-n$. This is essential for applications like live sports commentary, real-time video surveillance for anomaly detection, or accessibility tools that provide live descriptions of the visual world for the visually impaired. By maintaining an evolving memory of the visual stream, the model can provide coherent, context-rich interpretations without the need to process the entire video offline.

\subsection{Research Scope and Positioning}
This work positions itself as a pioneering exploration into the \textbf{paradigm of Streaming Thinking}—technically enabling Large Language Models (LLMs) to perform concurrent reading and reasoning. Our primary contribution lies in establishing the architectural and training methodologies required to adapt LLMs for efficient, continuous data processing.

Regarding our evaluation scope, we utilize mathematical reasoning, logic reasoning, and contextual QA tasks primarily as a controlled testbed to verify the logical consistency of our streaming framework. These deterministic domains serve as a rigorous testbed to verify that our model can maintain logical consistency and accuracy under the strict constraints of streaming inference. Our future work aims to extend this paradigm to complex streaming scenarios.

\section{Streaming Thinking Paradigm}
\label{Appendeix:Paradigm}

\subsection{Cognitive Foundations of Streaming Thinking}

Human reasoning during reading naturally unfolds in a streaming manner—people engage in comprehension and inference as information arrives, without waiting for the complete context.

Our proposed StreamingThinker architecture draws structural inspiration from established models of human discourse comprehension, specifically the Construction-Integration (CI) model proposed by Kintsch~\citep{kintsch1988role}. The CI model posits that comprehension occurs in two alternating cycles:
\begin{itemize}
    \item The Construction Phase: A bottom-up process where linguistic input triggers the activation of concepts and propositions based on local context. In this stage, the cognitive system acts as a high-bandwidth information buffer, prioritizing the establishment of local coherence (e.g., resolving immediate pronouns or connecting adjacent clauses) over global consistency. Crucially, this phase is non-selective: it allows multiple, potentially conflicting interpretations to co-exist in a temporary cognitive state. This ensures that no critical information is prematurely discarded before the full context is available

    \item The Integration Phase: Once the initial propositional network is constructed, the cognitive system iteratively updates node activations based on their connectivity and connection strengths. Through this process, the system resolves local ambiguities (e.g., polysemy) and filters out contextually inappropriate inferences. The result is a refined macrostructure where only the information strictly consistent with the global context remains active, ensuring a unified and logical understanding of the discourse.
\end{itemize}

Inspired by this cognitive process, we proposed in the main text a Streaming Thinking Paradigm that enables large language models to reason concurrently with incremental input.
Our streaming phase corresponds to the construction phase, where the model performs "shallow reasoning" (e.g., entity tracking, intermediate calculation) to process incoming chunks and maintain local coherence~\citep{graesser1994constructing}. Our final reasoning phase corresponds to the integration phase, where the model utilizes the fully accumulated context (KV cache) to synthesize a globally consistent answer. This theoretical alignment explains why our model avoids the "hallucination" pitfalls of premature guessing: like a human reader, it defers the final integration of complex causal chains until the necessary global information is available.

In this appendix, we further elaborate on this paradigm by providing detailed explanations and illustrative examples that clarify its operational design and the ordering of context and question.

At each step, the model incrementally processes the incoming sentence, focusing on shallow and progressive comprehension, which aligns the construction phase in human comprehension.
Specifically, we designed distinct categories of tasks for the model during local construction, serving as cognitive scaffolds for establishing local coherence. These tasks (e.g., intermediate calculation, entity tracking) compel the model to explicitly process and encode immediate details, transforming raw input into structured representations without prematurely committing to a global conclusion.
(1) \hlgreen{understanding and summarizing key information,}
(2) \hlpink{explaining ambiguities and reorganizing semantic relations,}
(3) \hlgray{extending logical implications,} and
(4) \hlorange{skipping thinking step when the content is irrelevant to the question.}

After completing incrementally reading and reasoning, the model shifts its focus from local coherence to global interpretation. It leverages the full context now available in its memory (KV cache) to perform the deep, unconstrained reasoning that was deliberately deferred during the streaming phase.
Specifically, the model may:
(1) directly produce an answer, representing the shallowest depth; (2) further integrate global information to achieve deeper comprehension; or (3) perform reflective reasoning on top of global integration to obtain the most reliable reasoning outcome.

\subsection{Examples of Paradigm Design}
We instantiate Streaming Thinking on three representative tasks---math reasoning, logical reasoning, and long-context QA---to show how the paradigm incrementally processes input, manages intermediate reasoning, and enables depth-wise answers (see Box B.1a--c).

\paragraph{Math reasoning (Box B.1a).}
Given a multi-conditions problem, the model reads the context sentence by sentence, identifying rate/quantity information, skipping irrelevant text, and maintaining interpretable intermediate results (e.g., partial counts). The pipeline supports three reasoning depths: D1 a direct computation from the current notes, D2 a globally consolidated solution that aggregates all intervals, and D3 a reflective pass that re-checks each arithmetic step to ensure reliability.
\begin{tcolorbox}[
    title=Paradigm design for Streaming Thinking in Math Reasoning,
    breakable,
    colback=white,
    width=\textwidth,
    boxrule=1pt,
    colframe=gray,
    label={box:math_reasoning}
]
\begin{minipage}[t]{0.49\textwidth}

\begin{center}
    \textbf{Input sentences}
\end{center}

\fontsize{6pt}{10pt}\selectfont

\textbf{Question}\\
\hlpink{\textbf{1.} How many more popcorns does Mohammed eat than his brother?}

\textbf{Context}\\
\hlorange{\textbf{2.} Mohammed is popping popcorn for a snack.}\\
\hlgreen{\textbf{3.} As the skillet of kernels heats up, the kernels start popping faster.}\\
\hlgray{\textbf{4.} 100 pop in the first 22 seconds of cooking, then 3 times that amount in the next 22 seconds.}\\

\hlgray{\textbf{5.} The kernels increase to 6 times the initial popping rate in the next 22 seconds.}\\

\hlgray{\textbf{6.} In the final 22 seconds, the popping slows down to half the rate as the past 22 seconds.}\\
\hlgray{\textbf{7.} After Mohammed takes the skillet off the heat, a quarter of the number of kernels that popped in the final 22 seconds of cooking also pop from the residual heat.}\\
\hlgray{\textbf{8.} Mohammed gives 2/5 of the made popcorn to his brother Yusuf.}\\

\end{minipage}\hfill
\begin{minipage}[t]{0.48\textwidth}
\begin{center}
    \textbf{Streaming thinking}
\end{center}

\fontsize{6pt}{10pt}\selectfont

\texttt{<think>}\\
\hlpink{\textbf{1.} Okay, I need to calculate how many more pieces of popcorn Mohammed has than his brother.}\\

\hlorange{\textbf{2.} \texttt{<Skip>}}\\
\hlgreen{\textbf{3.} Heating the skillet accelerates kernel popping.}\\
\hlgray{\textbf{4.} The process starts with 100 kernels popping in the first 22-second interval, followed by a second interval where the amount triples, yielding 100×3=300 kernels.}\\
\hlgray{\textbf{5.} In the next 22 seconds, the rate increases to 6 times the initial rate, meaning the number of popped kernels is based on the initial amount of 100, resulting in 100×6=600 kernels for this third period.}\\
\hlgray{\textbf{6.} For the final 22 seconds of cooking, the rate slows to half of the preceding interval's rate, which gives us 600÷2=300 kernels.}\\
\hlgray{\textbf{7.} After the skillet is removed from the heat, residual heat pops an additional quarter of the amount from the final cooking period, adding 300÷4=75 kernels.}\\
\hlgray{\textbf{8.} Finally, the total amount of popcorn is the sum of all intervals: 100+300+600+300+75=1375. Mohammed gives 2/5 of this to his brother Yusuf, which amounts to 1375×2÷5=550 kernels.}\\
\rule{\linewidth}{0.1pt}
\vspace{0.1pt}
[\textbf{\textit{Optional depth, streaming thinking with direct answer}}]\\
\hlblue{So, the answer is 275.}

[\textbf{{\textit{Optional depth, streaming thinking with global thinking}}}]\\
\hlblue{So, after giving 550 pieces to his brother, Mohammed retains 1375-550=825. The difference is 825-550=275.}

[\textbf{{\textit{Optional depth, streaming thinking with global thinking \& reflection}}}]\\
\hlblue{So, after giving 550 pieces to his brother, Mohammed retains 1375-550=825. The difference is 825-550=275. \textbf{Wait, let me check again.} The first interval is 100, second is 3× that = 300, third is 6× the initial rate (not cumulative), so 600—that's right. Fourth is half of the previous interval's rate, so half of 600 is 300, yes. Then residual heat adds a quarter of that final 300, so 75—correct. Total: 100+300+600+300+75=1,375. He gives 2/5 to Yusuf: 1,375×2÷5=550. That leaves Mohammed with 825. Difference: 825-550=275. All steps check out. So yes, Mohammed has 275 more pieces than his brother.}

\texttt{<\textbackslash think>}
\end{minipage}
\end{tcolorbox}

\paragraph{Logical reasoning (Box B.1b).}

Premises are normalized as explicit statements; each incoming sentence is mapped to entities and relations, while unrelated sentences are skipped. Local implications are updated online, and the final decision is produced at different depths: D1 yields an immediate verdict when premises suffice; D2 integrates all premises to assess entailment vs. contradiction; D3 performs a reflective consistency check, which may revise a premature ``True/False'' to ``Unknown'' when a critical premise (e.g., a location) is not actually supported by the context.

\begin{tcolorbox}[
    title=Paradigm design for Streaming Thinking in Logic Reasoning,
    breakable,
    colback=white,
    width=\textwidth,
    boxrule=1pt,
    colframe=gray
]
\begin{minipage}[t]{0.49\textwidth}

\begin{center}
    \textbf{Input sentences}
\end{center}

\fontsize{6pt}{10pt}\selectfont

\textbf{Question}\\
\hlgray{\textbf{1.} Is the conclusion true given the premises?, ['True', 'False', 'Unknown'], conclusion:William Dickinson went to schools located in the United Kingdom for both high school and university.}

\textbf{Context}\\
\hlpink{\textbf{2.}William Dickinson was a British politician who sat in the House of Commons William Dickinson attended Westminster school for high school and then the University of Edinburgh.}\\
\hlgreen{\textbf{3.} The University of Edinburgh is a university located in the United Kingdom.}\\
\hlorange{\textbf{4.} William Dickinson supported the Portland Whigs.}\\
\hlorange{\textbf{5.} People who supported the Portland Whigs did not get a seat in the Parliament.}\\

\end{minipage}\hfill
\begin{minipage}[t]{0.48\textwidth}
\begin{center}
    \textbf{Streaming thinking}
\end{center}

\fontsize{6pt}{10pt}\selectfont

\texttt{<think>}\\
\hlgray{\textbf{1.} Okay, the question is asking whether the conclusion — that William Dickinson went to schools in the United Kingdom for both high school and university — is true, based on the given premises. The possible answers are True, False, or Unknown. }\\
\hlpink{\textbf{2.} William Dickinson attended Westminster School for high school and then the University of Edinburgh.}\\

\hlgreen{\textbf{3.} The University of Edinburgh is located in the United Kingdom.}\\

\hlorange{\textbf{4.} \texttt{<Skip>}}\\
\hlorange{\textbf{5.} \texttt{<Skip>}}\\

\rule{\linewidth}{0.1pt}
\vspace{0.1pt}
[\textbf{\textit{Optional depth, streaming thinking with direct answer}}]\\
\hlblue{Therefore, the answer is True.}

[\textbf{{\textit{Optional depth, streaming thinking with global thinking}}}]\\
\hlblue{The conclusion specifies that both his high school and university were located in the United Kingdom. The context explicitly states he attended Westminster School and the University of Edinburgh. Since the University of Edinburgh is confirmed to be in the UK, the university part is satisfied. However, the location of Westminster School is not directly stated in the context. Westminster School is a well-known school in London, UK, but the context does not explicitly mention its location. Without explicit confirmation that Westminster School is in the UK, the conclusion about both institutions being in the UK remains uncertain. Therefore, the answer is Unknown.}

[\textbf{{\textit{Optional depth, streaming thinking with global thinking \& reflection}}}]\\
\hlblue{The conclusion specifies that both his high school and university were located in the United Kingdom. The context explicitly states he attended Westminster School and the University of Edinburgh. Since the University of Edinburgh is confirmed to be in the UK, the university part is satisfied. However, the location of Westminster School is not directly stated in the context. Westminster School is a well-known school in London, UK, but the context does not explicitly mention its location. Without explicit confirmation that Westminster School is in the UK, the conclusion about both institutions being in the UK remains uncertain. \textbf{Wait, let me check again.} The context only mentions the University of Edinburgh's location in the UK. Westminster School is not explicitly stated to be in the UK. Even though common knowledge might suggest it is, the premises provided do not confirm this. Therefore, the conclusion cannot be fully verified from the given premises. Therefore, the answer is Unknown.}

\texttt{<\textbackslash think>}
\end{minipage}
\end{tcolorbox}

\paragraph{Context-based QA (Box B.1c).}
For scientific or document-level passages, the model streams through long context, filtering sentences that are irrelevant, extracting mechanistic cues (e.g., causal roles, experimental outcomes), and merging compatible evidence across distant spans. D1 gives a concise answer when a single decisive statement appears; D2 synthesizes multi-sentence evidence for a stronger justification; D3 runs a reflective verification that cross-checks all cited evidence for agreement before committing to the final answer.

\begin{tcolorbox}[
    title=Paradigm Design for Streaming Thinking in Context-based QA Reasoning,
    breakable,
    colback=white,
    width=\textwidth,
    boxrule=1pt,
    colframe=gray,
    label={box:QA_reasoning}
]
\begin{minipage}[t]{0.49\textwidth}

\begin{center}
    \textbf{Input sentences}
\end{center}

\fontsize{6pt}{10pt}\selectfont

\textbf{Question}\\
\hlpink{\textbf{1.} Do mitochondria play a role in remodelling lace plant leaves during programmed cell death?}

\textbf{Context}\\
\hlpink{\textbf{2.} Programmed cell death (PCD) is the regulated death of cells within an organism.}\\
\hlpink{\textbf{3.} The lace plant (Aponogeton madagascariensis) produces perforations in its leaves through PCD.}\\
\hlorange{\textbf{4.} The leaves of the plant consist of a latticework of longitudinal and transverse veins enclosing areoles.}\\
\hlorange{\textbf{5.} PCD occurs in the cells at the center of these areoles and progresses outwards, stopping approximately five cells from the vasculature.}\\
\hlgreen{\textbf{6.} The role of mitochondria during PCD has been recognized in animals; however, it has been less studied during PCD in plants.}\\
\hlpink{\textbf{7.} The following paper elucidates the role of mitochondrial dynamics during developmentally regulated PCD in vivo in A. madagascariensis. A single areole within a window stage leaf (PCD is occurring) was divided into three areas based on the progression of PCD; cells that will not undergo PCD (NPCD), cells in early stages of PCD (EPCD), and cells in late stages of PCD (LPCD).}\\
\hlpink{\textbf{8.} Window stage leaves were stained with the mitochondrial dye MitoTracker Red CMXRos and examined.}\\
\hlpink{\textbf{9.} Mitochondrial dynamics were delineated into four categories (M1-M4) based on characteristics including distribution, motility, and membrane potential ($\Delta \Psi_m$).}\\
\hlgray{\textbf{10.} A TUNEL assay showed fragmented nDNA in a gradient over these mitochondrial stages.}\\
\hlorange{\textbf{11.} Chloroplasts and transvacuolar strands were also examined using live cell imaging.}\\
\hlpink{\textbf{12.} The possible importance of mitochondrial permeability transition pore (PTP) formation during PCD was indirectly examined via in vivo cyclosporine A (CsA) treatment.}\\
\hlgray{\textbf{13.} This treatment resulted in lace plant leaves with a significantly lower number of perforations compared to controls, and that displayed mitochondrial dynamics similar to that of non-PCD cells.}\\

\end{minipage}\hfill
\begin{minipage}[t]{0.48\textwidth}
\begin{center}
    \textbf{Streaming thinking}
\end{center}

\fontsize{6pt}{10pt}\selectfont

\texttt{<think>}\\
\hlpink{\textbf{1.} Okay, let's see. The question asks whether mitochondria contribute to the remodeling of lace plant leaves during programmed cell death. I need to focus on 'mitochondria', 'remodeling of lace plant leaves', and 'programmed cell death'.}\\
\hlpink{\textbf{2.} The first sentence introduce that Programmed cell death (PCD) is the regulated death of cells within an organism.}\\
\hlpink{\textbf{3.} In lace plants, leaf perforations are created through programmed cell death.}\\
\hlorange{\textbf{4.} \texttt{<Skip>}}\\
\hlorange{\textbf{5.} \texttt{<Skip>}}\\
\hlgreen{\textbf{6.} This study investigates mitochondrial involvement in developmental cell death in lace plant leaves.}\\
\hlpink{\textbf{7.} Researchers divided an areole into three zones: non-PCD cells, early PCD cells, and late PCD cells, reflecting the progression of cell death.}\\
\hlpink{\textbf{8.} They stained mitochondria in window-stage leaves with MitoTracker Red to observe their properties.}\\
\hlpink{\textbf{9.} Mitochondria were classified into four dynamic states (M1–M4) based on distribution, motility, and membrane potential.}\\
\hlgray{\textbf{10.} A DNA fragmentation assay (TUNEL) showed nuclear degradation corresponding with mitochondrial changes, linking mitochondrial dynamics to PCD progression.}\\
\hlorange{\textbf{11.} \texttt{<Skip>}}\\
\hlpink{\textbf{12.} The potential involvement of the mitochondrial permeability transition pore was tested pharmacologically with cyclosporine A.}\\
\hlgray{\textbf{13.} Treatment reduced the number of leaf perforations and preserved mitochondria in a state similar to healthy cells, showing that blocking mitochondrial pore activity suppressed PCD.}\\
\rule{\linewidth}{0.1pt}
\vspace{0.1pt}
[\textbf{\textit{Optional depth, streaming thinking with direct answer}}]\\
\hlblue{Therefore, the answer is No.}

[\textbf{{\textit{Optional depth, streaming thinking with global thinking}}}]\\
\hlblue{The study demonstrates that mitochondria change in distribution, motility, and membrane potential during the stages of programmed cell death in lace plant leaves. DNA fragmentation corresponds with these mitochondrial states, and inhibition of the mitochondrial permeability transition pore reduces perforation formation. Together, this evidence shows that mitochondria are actively involved in regulating and executing cell death that sculpts lace plant leaves. Therefore, the answer is Yes.}

[\textbf{{\textit{Optional depth, streaming thinking with global thinking \& reflection}}}]\\
\hlblue{The study demonstrates that mitochondria change in distribution, motility, and membrane potential during the stages of programmed cell death in lace plant leaves. DNA fragmentation corresponds with these mitochondrial states, and inhibition of the mitochondrial permeability transition pore reduces perforation formation. Together, this evidence shows that mitochondria are actively involved in regulating and executing cell death that sculpts lace plant leaves. \textbf{Wait, let me check again.} The lace plant forms leaf perforations through PCD, so if mitochondria affect PCD, they affect remodeling. The study showed mitochondria shift in distribution, motility, and membrane potential during PCD stages, changes that align with DNA fragmentation. Blocking the mitochondrial permeability transition pore reduced perforations and preserved healthy-like mitochondria. Thus, mitochondria actively regulate and execute PCD in shaping lace plant leaves. Therefore, the answer is Yes.}

\texttt{<\textbackslash think>}
\end{minipage}
\end{tcolorbox}

\newpage
\subsection{Examples of the Ordering of Context and Question}

\vspace{-0.5em}
\paragraph{Math reasoning (Box B.2a).}
\emph{Question-first} (left panel) anchors the target quantity upfront, so each incoming sentence is parsed for rates/amounts relative to the goal, with irrelevant lines skipped and partial counts maintained online. \emph{Context-first} (right panel) encourages incremental accumulation of numerical evidence without a declared target; once the question appears, the model consolidates cached quantities to compute the final difference. Both settings yield interpretable traces but emphasize goal-driven vs.\ evidence-driven processing, respectively.
\vspace{-0.5em}
\vfill
\begin{tcolorbox}[
    title=Example: Order of Context and Question for Streaming Thinking in Math Reasoning,
    breakable,
    colback=white,
    width=\textwidth,
    boxrule=1pt,
    colframe=gray
]
\begin{minipage}[t]{0.49\textwidth}

\textbf{Question first}
\fontsize{6pt}{10pt}\selectfont
\vspace{-1em}
\begin{center}
    \textbf{Input sentences}
\end{center}
\vspace{-1em}

\textbf{Question}\\
\textbf{1.} How many more popcorns does Mohammed eat than his brother?

\textbf{Context}\\
\textbf{2.} Mohammed is popping popcorn for a snack.\\
\textbf{3.} As the skillet of kernels heats up, the kernels start popping faster.\\
\textbf{4.} 100 pop in the first 22 seconds of cooking, then 3 times that amount in the next 22 seconds.\\
\textbf{5.} The kernels increase to 6 times the initial popping rate in the next 22 s.\\
\textbf{6.} In the final 22 seconds, the popping slows down to half the rate as the past 22 seconds.\\
\textbf{7.} After Mohammed takes the skillet off the heat, a quarter of the number of kernels that popped in the final 22 seconds of cooking also pop from the residual heat.\\
\textbf{8.} Mohammed gives 2/5 of the made popcorn to his brother Yusuf.

\vspace{-1em}
\begin{center}
    \textbf{Streaming thinking}
\end{center}
\fontsize{6pt}{10pt}\selectfont
\vspace{-1em}

\texttt{<think>}\\
\hlpink{\textbf{1.} Okay, I need to calculate how many more pieces of popcorn Mohammed has than his brother.}\\
\hlorange{\textbf{2.} \texttt{<Skip>}}\\
\textbf{3.} Heating the skillet accelerates kernel popping.\\
\textbf{4.} The process starts with 100 kernels popping in the first 22-second interval, followed by a second interval where the amount triples, yielding 100×3=300 kernels.\\\
\textbf{5.} In the next 22 seconds, the rate increases to 6 times the initial rate, meaning the number of popped kernels is based on the initial amount of 100, resulting in 100×6=600 kernels for this third period.\\
\textbf{6.} For the final 22 seconds of cooking, the rate slows to half of the preceding interval's rate, which gives us 600÷2=300 kernels.\\
\textbf{7.} After the skillet is removed from the heat, residual heat pops an additional quarter of the amount from the final cooking period, adding 300÷4=75 kernels.\\
\textbf{8.} Finally, the total amount of popcorn is the sum of all intervals: 100+300+600+300+75=1375. Mohammed gives 2/5 of this to his brother Yusuf, which amounts to 1375×2÷5=550 kernels.\\
\rule{\linewidth}{0.4pt}
[\textbf{\textit{Optional depth, streaming thinking with direct answer}}]\\
So, the answer is 275.

[\textbf{{\textit{Optional depth, streaming thinking with global thinking}}}]\\
So, after giving 550 pieces to his brother, Mohammed retains 1375-550=825. The difference is 825-550=275.

[\textbf{{\textit{Optional depth, streaming thinking with global thinking \& reflection}}}]\\
So, after giving 550 pieces to his brother, Mohammed retains 1375-550=825. The difference is 825-550=275. \textbf{Wait, let me check again.} The first interval is 100, second is 3× that = 300, third is 6× the initial rate (not cumulative), so 600—that's right. Fourth is half of the previous interval's rate, so half of 600 is 300, yes. Then residual heat adds a quarter of that final 300, so 75—correct. Total: 100+300+600+300+75=1,375. He gives 2/5 to Yusuf: 1,375×2÷5=550. That leaves Mohammed with 825. Difference: 825-550=275. All steps check out. So yes, Mohammed has 275 more pieces than his brother.

\texttt{<\textbackslash think>}

\end{minipage}\hfill
\begin{minipage}[t]{0.49\textwidth}

\textbf{Context first}
\fontsize{6pt}{10pt}\selectfont
\vspace{-1em}
\begin{center}
    \textbf{Input sentences}
\end{center}
\vspace{-1em}

\textbf{Context}\\
{\textbf{1.} Mohammed is popping popcorn for a snack.}\\
{\textbf{2.} As the skillet of kernels heats up, the kernels start popping faster.}\\
{\textbf{3.} 100 pop in the first 22 seconds of cooking, then 3 times that amount in the next 22 seconds.}\\
\textbf{4.} The kernels increase to 6 times the initial popping rate in the next 22 s.\\
\textbf{5.} In the final 22 seconds, the popping slows down to half the rate as the past 22 seconds.\\
\textbf{6.} After Mohammed takes the skillet off the heat, a quarter of the number of kernels that popped in the final 22 seconds of cooking also pop from the residual heat.\\
\textbf{7.} Mohammed gives 2/5 of the made popcorn to his brother Yusuf.\\
\textbf{Question}\\
\textbf{8.} How many more popcorns does Mohammed eat than his brother?

\vspace{-1em}
\begin{center}
    \textbf{Streaming thinking}
\end{center}
\fontsize{6pt}{10pt}\selectfont
\vspace{-1em}

\texttt{<think>}\\
\hlorange{\textbf{1.} Okay, the input sentences are start with that Mohammed pops popcorn for a snack.}\\
{\textbf{2.} Then we know that popping speeds up as the skillet heats.}\\
\textbf{3.} Then the process starts with 100 kernels popping in the first 22-second interval, followed by a second interval where the amount triples, yielding 100×3=300 kernels.\\
\textbf{4.} In the next 22 seconds, the rate increases to 6 times the initial rate, meaning the number of popped kernels is based on the initial amount of 100, resulting in 100×6=600 kernels for this third period.\\
\textbf{5.} For the final 22 seconds of cooking, the rate slows to half of the preceding interval's rate, which gives us 600÷2=300 kernels.\\
\textbf{6.} After the skillet is removed from the heat, residual heat pops an additional quarter of the amount from the final cooking period, adding 300÷4=75 kernels.\\
\textbf{7.} Finally, the total amount of popcorn is the sum of all intervals: 100+300+600+300+75=1375. Mohammed gives 2/5 of this to his brother Yusuf, which amounts to 1375×2÷5=550 kernels.\\
\hlpink{\textbf{8.} Now I have read the question and I need to calculate how many more pieces of popcorn Mohammed has than his brother.}\\
\rule{\linewidth}{0.4pt}
[\textbf{\textit{Optional depth, streaming thinking with direct answer}}]\\
So, the answer is 275.

[\textbf{\textit{Optional depth, streaming thinking with global thinking}}]\\
So, after giving 550 pieces to his brother, Mohammed retains 1375-550=825. The difference is 825-550=275.

[\textbf{\textit{Optional depth, streaming thinking with global thinking \& reflection}}]\\
So, after giving 550 pieces to his brother, Mohammed retains 1375-550=825. The difference is 825-550=275. \textbf{Wait, let me check again.} The first interval was 100, then three times that → 300, then six times the initial rate (not cumulative) → 600. After that, half of the previous 600 → 300, and finally the residual quarter of that 300 → 75. Adding them all, 100 + 300 + 600 + 300 + 75 = 1375, consistent. Giving 2/5 to Yusuf: 1375 × 2 ÷ 5 = 550. Subtracting leaves 825, and the difference indeed comes out as 275. So yes, after rechecking, the conclusion holds—Mohammed ends up with 275 more pieces than his brother.

\texttt{<\textbackslash think>}
\vspace{-1em}
\end{minipage}
\end{tcolorbox}

\paragraph{Logical reasoning (Box B.2b).}
Under \emph{question-first} ordering, the model enters a hypothesis-testing mode: each premise is checked against the claimed conclusion, distractors are skipped, and if a critical premise is missing, depth D3 returns \textsc{Unknown}. With \emph{context-first} ordering, the model first normalizes premises into entities and relations; after the question is posed, it adjudicates entailment/contradiction/unknown using the cached structure. Answers remain consistent across orders, while the traces reveal different timing of inference.

\vfill
\begin{tcolorbox}[
    title=Example: Order of Context and Question for Streaming Thinking in Logical Reasoning,
    breakable,
    colback=white,
    width=\textwidth,
    boxrule=1pt,
    colframe=gray
]
\begin{minipage}[t]{0.49\textwidth}

\textbf{Question first}
\fontsize{6pt}{10pt}\selectfont
\begin{center}
    \textbf{Input sentences}
\end{center}

\textbf{Question}\\
\textbf{1.} Is the conclusion true given the premises?, ['True', 'False', 'Unknown'], conclusion:William Dickinson went to schools located in the United Kingdom for both high school and university.

\textbf{Context}\\
\textbf{2.} William Dickinson was a British politician who sat in the House of Commons William Dickinson attended Westminster school for high school and then the University of Edinburgh.\\
\textbf{3.} The University of Edinburgh is a university located in the UK.\\
\textbf{4.} William Dickinson supported the Portland Whigs.\\
\textbf{5.} People who supported the Portland Whigs did not get a seat in the Parliament.

\vspace{-1em}
\begin{center}
    \textbf{Streaming thinking}
\end{center}
\fontsize{6pt}{10pt}\selectfont
\vspace{-1em}

\texttt{<think>}\\
\hlpink{\textbf{1.} Okay, the question is asking whether the conclusion — that William Dickinson went to schools in the United Kingdom for both high school and university — is true, based on the given premises. The possible answers are True, False, or Unknown.}\\
\hlorange{\textbf{2.} William Dickinson attended Westminster School for high school and then the University of Edinburgh.}\\
\textbf{3.} The University of Edinburgh is located in the United Kingdom.\\
\textbf{4.} \texttt{<Skip>}\\
\textbf{5.} \texttt{<Skip>}\\
\rule{\linewidth}{0.4pt}
[\textbf{\textit{Optional depth, streaming thinking with direct answer}}]\\
So, the answer is 275.

[\textbf{{\textit{Optional depth, streaming thinking with global thinking}}}]\\
The conclusion specifies that both his high school and university were located in the UK. The context explicitly states he attended Westminster School and the University of Edinburgh. Since the University of Edinburgh is confirmed to be in the UK, the university part is satisfied. However, the location of Westminster School is not directly stated in the context. Westminster School is a well-known school in London, UK, but the context does not explicitly mention its location. Without explicit confirmation that Westminster School is in the UK, the conclusion about both institutions being in the UK remains uncertain. Therefore, the answer is Unknown.

[\textbf{{\textit{Optional depth, streaming thinking with global thinking \& reflection}}}]\\
The conclusion specifies that both his high school and university were located in the United Kingdom. The context explicitly states he attended Westminster School and the University of Edinburgh. Since the University of Edinburgh is confirmed to be in the UK, the university part is satisfied. However, the location of Westminster School is not directly stated in the context. Westminster School is a well-known school in London, UK, but the context does not explicitly mention its location. Without explicit confirmation that Westminster School is in the UK, the conclusion about both institutions being in the UK remains uncertain. Wait, let me check again. The context only mentions the University of Edinburgh's location in the UK. Westminster School is not explicitly stated to be in the UK. Even though common knowledge might suggest it is, the premises provided do not confirm this. Therefore, the conclusion cannot be fully verified from the given premises. Therefore, the answer is Unknown.

\texttt{<\textbackslash think>}
\vspace{-1em}

\end{minipage}\hfill
\begin{minipage}[t]{0.49\textwidth}

\textbf{Context first}
\fontsize{6pt}{10pt}\selectfont
\begin{center}
    \textbf{Input sentences}
\end{center}

\textbf{Context}\\
{\textbf{1.} William Dickinson was a British politician who sat in the House of Commons William Dickinson attended Westminster school for high school and then the University of Edinburgh.}\\
{\textbf{2.} The University of Edinburgh is a university located in the UK.}\\
{\textbf{3.} William Dickinson supported the Portland Whigs.}\\
\textbf{4.} People who supported the Portland Whigs did not get a seat in the Parliament. \\
\textbf{Question}\\
\textbf{5.} Is the conclusion true given the premises?, ['True', 'False', 'Unknown'], conclusion:William Dickinson went to schools located in the United Kingdom for both high school and university.

\vspace{-1em}
\begin{center}
    \textbf{Streaming thinking}
\end{center}
\fontsize{6pt}{10pt}\selectfont
\vspace{-1em}

\texttt{<think>}\\
\hlorange{\textbf{1.} Okay, we now known that William Dickinson attended Westminster School for high school and then the University of Edinburgh.}\\
{\textbf{2.} And, the University of Edinburgh is a university situated in the United Kingdom.}\\
\textbf{3.} William Dickinson supports the Portland Whigs.\\
\textbf{4.} Then, according to the rule, anyone who supported the Portland Whigs did not obtain a seat in Parliament. So William Dickinson did not obtain a seat in Parliament.\\
\hlpink{\textbf{5.} Now, the question is asking whether the conclusion — that William Dickinson went to schools in the United Kingdom for both high school and university — is true.}\\
\rule{\linewidth}{0.4pt}
[\textbf{\textit{Optional depth, streaming thinking with direct answer}}]\\
Therefore, the answer is True.

[\textbf{\textit{Optional depth, streaming thinking with global thinking}}]\\
The context explicitly states he attended Westminster School and the University of Edinburgh. Since the University of Edinburgh is confirmed to be in the UK, the university part is satisfied. However, the location of Westminster School is not directly stated in the context. Westminster School is a well-known school in London, UK, but the context does not explicitly mention its location. Without explicit confirmation that Westminster School is in the UK, the conclusion about both institutions being in the UK remains uncertain. Therefore, the answer is Unknown.

[\textbf{\textit{Optional depth, streaming thinking with global thinking \& reflection}}]\\
The context explicitly states he attended Westminster School and the University of Edinburgh. Since the University of Edinburgh is confirmed to be in the UK, the university part is satisfied. However, the location of Westminster School is not directly stated in the context. Westminster School is a well-known school in London, UK, but the context does not explicitly mention its location. Without explicit confirmation that Westminster School is in the UK, the conclusion about both institutions being in the UK remains uncertain. Wait, let me check again. The context only mentions the University of Edinburgh's location in the UK. Westminster School is not explicitly stated to be in the UK. Even though common knowledge might suggest it is, the premises provided do not confirm this. Therefore, the conclusion cannot be fully verified from the given premises. Therefore, the answer is Unknown.

\texttt{<\textbackslash think>}
\vspace{-1em}
\end{minipage}
\end{tcolorbox}

\newpage
\section{Streaming CoT Generation}
\label{Appendeix:Generation}
\subsection{Does Incremental Sentence Input Work?}
A straightforward idea is to feed sentences into LLMs sequentially and expect sentence-by-sentence reasoning based on the accumulated input. However, our experiments show that when given only a single sentence without streaming fine-tuning, the original reasoning model tends to overthink~\citep{sui2025stop}. In other words, although this sentence-level data generation method appears to align better with the streaming property, the vanilla model lacks such generative capability and exhibits reduced controllability in CoT generation.

\subsection{Does Direct Prompting of LLMs Work?}
Another intuitive approach is to directly prompt LLMs to perform streaming-style reasoning with carefully crafted instructions. In this setup, the model is explicitly asked to reason sentence by sentence. Nevertheless, our experiments show that, lacking fine-grained control, prompt engineering alone is insufficient: although the model may superficially follow the instructions, it often deviates from the intended trajectory, resulting in inconsistent or overly verbose reasoning.
Therefore, we design a pipeline for CoT trajectory control based on the complete input, which enforces sequential reasoning in LLMs through explicit boundary constraints and teacher guidance.

\subsection{Control CoT Process}
In this work, we design a multi-stage pipeline for streaming CoT generation and control. The pipeline consists of four key components: (1) boundary-token insertion to finely define reasoning units and ensure sequential alignment; (2) instruction-based prompting to guide the LLM toward the desired streaming paradigm; (3) teacher-model guidance to reconstruct reasoning chains with improved structural consistency and semantic fidelity; and (4) token-level intervention to control the trajectory of CoT, enabling adjustable reasoning depth and style.
\paragraph{Boundary Token Insert}
In our proposed streaming thinking paradigm, a complete sentence serves as a atomic unit of cognition. This design achieves a critical balance between the fine-grained nature of streaming processing and the preservation of semantic integrity in the information being processed. To implement this design, we introduce a structured protocol of boundary tokens to explicitly demarcate the model's thinking boundaries. Specifically, this protocol distinguishes between input signals and the model's generated reasoning states:
\begin{itemize}
    \item
    Input Demarcation: We employ two distinct tokens to structure the input stream. The \texttt{<EOS>} (End-of-Sentence) token follows each contextual sentence, triggering an incremental thinking step for context assimilation. In contrast, the \texttt{<EOQ>} (End-of-Question) token marks the end of the problem description, signaling a shift from context processing to answer formulation.
    \item
    Thinking State Demarcation: The model's thought process is, in turn, segmented by three corresponding output tokens. The \texttt{<EOT>} (End-of-Thought) token concludes each incremental reasoning segment associated with a context sentence. The \texttt{<EOQ>} token concludes the reasoning phase triggered by the question. Finally, the \texttt{<EOR>} (End-of-Reasoning) token signifies the completion of the entire streaming thinking chain with controllable depth, preceding the final answer generation.
\end{itemize}
It is important to note that the boundary tokens are introduced primarily as a data generation and evaluation methodology. Their purpose is to provide a structured prompt for the LLMs to generate the desired streaming chain-of-thought behavior. Moreover, in real streaming reasoning scenarios, once the LLM outputs a boundary token, the current reasoning unit can be regarded as completed. Thus, during training, boundary tokens serve as supervisory signals that guide the model to recognize when a reasoning unit should be terminated.

There is an example for boundary tokens insert.
\begin{tcolorbox}[
    title=Boundary Tokens Insert for Generation,
    breakable,
    colback=white,
    width=\textwidth,
    boxrule=1pt,
    colframe=gray
]
\begin{minipage}[t]{0.49\textwidth}
\begin{center}
    \textbf{Input sentences}
\end{center}
\fontsize{6pt}{10pt}\selectfont
\vspace{-1em}

\textbf{Question}\\
1. How many more popcorns does Mohammed eat than his brother?\texttt{\textcolor{red}{<EOQ>}}

\textbf{Context}\\
{2. Mohammed is popping popcorn for a snack.\texttt{\textcolor{red}{<EOS>}}}\\
{3. As the skillet of kernels heats up, the kernels start popping faster.\texttt{\textcolor{red}{<EOS>}}}\\
{4. 100 pop in the first 22 seconds of cooking, then 3 times that amount in the next 22 seconds.\texttt{\textcolor{red}{<EOS>}}}\\

5. The kernels increase to 6 times the initial popping rate in the next 22 seconds.\texttt{\textcolor{red}{<EOS>}}\\

6. In the final 22 seconds, the popping slows down to half the rate as the past 22 seconds.\texttt{\textcolor{red}{<EOS>}}\\
7. After Mohammed takes the skillet off the heat, a quarter of the number of kernels that popped in the final 22 seconds of cooking also pop from the residual heat.\texttt{\textcolor{red}{<EOS>}}\\
8. Mohammed gives 2/5 of the made popcorn to his brother Yusuf.\texttt{\textcolor{red}{<EOS>}}\\

\end{minipage}\hfill
\begin{minipage}[t]{0.48\textwidth}
\begin{center}
    \textbf{Streaming thinking}
\end{center}
\vspace{-1em}

\fontsize{6pt}{10pt}\selectfont

\texttt{<think>}\\
1. Okay, I need to calculate how many more pieces of popcorn Mohammed has than his brother.\texttt{\textcolor{red}{<EOQ>}}\\
{2. \texttt{<Skip>\textcolor{red}{<EOT>}}}\\
{3. \texttt{<Skip>\textcolor{red}{<EOT>}}}\\
4. The process starts with 100 kernels popping in the first 22-second interval, followed by a second interval where the amount triples, yielding 100×3=300 kernels.\texttt{\textcolor{red}{<EOT>}}\\
5. In the next 22 seconds, the rate increases to 6 times the initial rate, meaning the number of popped kernels is based on the initial amount of 100, resulting in 100×6=600 kernels for this third period.\texttt{\textcolor{red}{<EOT>}}\\
6. For the final 22 seconds of cooking, the rate slows to half of the preceding interval's rate, which gives us 600÷2=300 kernels.\texttt{\textcolor{red}{<EOT>}}\\
7. After the skillet is removed from the heat, residual heat pops an additional quarter of the amount from the final cooking period, adding 300÷4=75 kernels.\texttt{\textcolor{red}{<EOT>}}\\
8. Finally, the total amount of popcorn is the sum of all intervals: 100+300+600+300+75=1375. Mohammed gives 2/5 of this to his brother Yusuf, which amounts to 1375×2÷5=550 kernels.\texttt{\textcolor{red}{<EOT>}}\\
\rule{\linewidth}{0.1pt}

So, after giving 550 pieces to his brother, Mohammed retains 1375-550=825. The difference is 825-550=275.\texttt{\textcolor{red}{<EOR>}}\par

\texttt{<\textbackslash think>}
\vspace{-1em}
\end{minipage}
\end{tcolorbox}

\paragraph{Prompt Template}

\vspace{-2em}
\begin{table}[h]
    \centering
    \footnotesize
    \caption{Streaming reasoning instruct template.}
    \label{Tab:Instruct_temp}
    \renewcommand{\arraystretch}{1.2}
    \begin{tabularx}{\linewidth}{X}
        \toprule
\textbf{Input:} \textcolor{red!60}{question, context, example.}\\
\midrule
You are a streaming reasoner to perform an order reasoning based on the following Question and Context:\\
Input Question:\texttt{\{\textcolor{red!80}{question}}\}\\
Input Context:\texttt{\{\textcolor{red!80}{context}}\}
\par\vspace{0.7em}
Follow these strict instructions:\\
1. Must process the Context in order, as if the input arrive in a streaming way.\\
(1) Must perform reasoning one sentence at a time, i.e. When you encounter an \texttt{<EOS>} in the Input Context, perform reasoning about the sentence that before this \texttt{<EOS>}.\\
(2) Upon encountering \texttt{<EOS>}, immediately reason about the preceding sentence and end with an \texttt{<EOT>}.\\
(3) Do Not skip any \texttt{<EOS>} and Do Not forget any \texttt{<EOT>}, i.e. each \texttt{<EOS>} in the Input must match a \texttt{<EOT>} in the reasoning.
\par\vspace{0.7em}
2. If you read any Question end with \texttt{<EOQ>}.\\
(1) Explain the Question that before the \texttt{<EOQ>} in the Input Question.
(2) End your interpretation with \texttt{<EOQ>}.
(3) Do not add assumptions beyond the explicit meaning of the Question.
\par\vspace{0.7em}
3. For each input sentence, smoothly integrate the following into a coherent paragraph-style output:\\
(1) \textbf{Understanding and summarizing key information}: (1-1) Ignore redundant, irrelevant, or ambiguous information. (1-2)Focus on extracting core facts, quantities, conditions, roles, time, location, and other essential elements.\\
(2) \textbf{Explaining ambiguities and reorganizing semantic relation}: (2-1) Clarify vague expressions, ambiguities, or implied information. (2-2) Rephrase the original meaning using clearer, more straightforward, and structurally explicit language, without changing the factual content.\\
(3) \textbf{Lightly extending logical implications}: (3-1) Allowed examples include: Simple calculations (e.g., totals, differences, ratios). Natural inferences about time, sequence, or spatial order. Direct factual inference that logically follows from explicit information. (3-2) Prohibited actions: assumptions, predictions, or subjective evaluations.\\
(4) \textbf{Skipping irrelevant input}: If a sentence is not related to the Question, output \texttt{<Skip><EOT>} immediately after reading \texttt{<EOS>}.
\par\vspace{0.7em}
4. \textbf{After reading and reasoning all sentencess: DO NOT summarize or infer any judgment.}
\par\vspace{0.7em}
There is an in-context example:\texttt{\{\textcolor{red!80}{example}}\}.\\

        \bottomrule
    \end{tabularx}
    \label{app_tab:prompt_template}
\end{table}

The prompt presented in~\autoref{app_tab:prompt_template} casts the model as a streaming reasoner that consumes the context sentence-by-sentence, keyed by \texttt{<EOS>}, and emits an immediate reasoning step ending with \texttt{<EOT>}. It first interprets any question ending with \texttt{<EOQ>} and closes that interpretation with \texttt{<EOQ>}. Each step must extract core facts, clarify ambiguities, and permit only light, local inferences; irrelevant sentences are output as \texttt{<Skip><EOT>}. No global summary is allowed, and an in-context example anchors the desired behavior.

\paragraph{Teacher Guidance Template}

The prompt shown in~\autoref{app_tab:prompt_teacher} directs a teacher LLM to realign an initial streaming trace so that every input sentence (terminated by \texttt{<EOS>}) maps exactly to one reasoning step (terminated by \texttt{<EOT>}). If present, the question is explained once up to \texttt{<EOQ>} before processing the context in strict order. The teacher may only simplify, clarify, and streamline existing content—no new reasoning, no cross-sentence references, no merges, skips, or delays—thereby enforcing strict locality and one-to-one pairing.
\begin{table}[h]
    \centering
    \footnotesize
    \caption{Instruction template for teacher LLM under both question-first and context-first settings.}
    \label{Tab:Reconstruct_temp}
    \renewcommand{\arraystretch}{1.2}
    \begin{tabularx}{\linewidth}{X}
        \toprule
\textbf{Input:} \textcolor{red!60}{question, context, example, reasoning.}\\
\midrule
You have to reconstruct the following reasoning process based on the Original Input and the Initial Streaming Reasoning Process, ensuring full compliance with the requirements of a streaming reasoner. The Original Input consists of multiple sentences, each ending with \texttt{<EOS>}. Every sentence represents an independent unit of reasoning. Your objective is to revise the initial reasoning such that each reasoning step corresponds exactly.

\par\vspace{0.7em}
Follow these strict instructions:\\
1. If any Question before reading the Input context.\\
(1) Explain the Question that before the \texttt{<EOQ>} in the Input Question at first.\\
(2) End your interpretation with \texttt{<EOQ>}.\\
(3) Do not add assumptions beyond the explicit meaning of the Question.
\par\vspace{0.7em}
2. Must process the Context in order, as if the input arrive in a streaming way.\\
(1) Must perform reasoning one sentence at a time, i.e. When you encounter an \texttt{<EOS>} in the Input Context, perform reasoning about the sentence that before this \texttt{<EOS>}.\\
(2) Upon encountering \texttt{<EOS>}, immediately reason about the preceding sentence and output a result ending with an \texttt{<EOT>}.\\
(3) Do Not skip any \texttt{<EOS>} and Do Not forget any \texttt{<EOT>}, i.e. each \texttt{<EOS>} in the Input must match a \texttt{<EOT>} in the reasoning.
\par\vspace{0.7em}
3. For each input sentence, smoothly integrate the following into a coherent paragraph-style reasoning output:\\
(1) {Understanding and summarizing key information}.\\
(2) {Explaining ambiguities and reorganizing semantic relation}.\\
(3) {Lightly extending logical implications}.\\
(4) {Skipping irrelevant input}.
\par\vspace{0.7em}
4. For each sentence:\\
(1) Do not merge, skip, or delay reasoning for any sentence. Each input sentence must be immediately followed by its reasoning step.\\
(2) Do not quote or repeat the input sentence in the reasoning output. \\
(3) Do not introduce new reasoning not found in the original streaming process. You may only simplify, clarify, or streamline what is already there.\\
(4) Your reasoning must be strictly local to each sentence. Do not refer to previous or later sentences. Avoid speculation, assumptions, or global summaries when streaming reasoning.
\par\vspace{0.7em}
There is an in-context example:\texttt{\{\textcolor{red!80}{example}}\}.
\par\vspace{0.7em}
Now, begin your streaming reasoning reconstruction:\\
Input question: \texttt{\{\textcolor{red!80}{question}\}}\\
Input context: \texttt{\{\textcolor{red!80}{context}\}}\\
Reasoning content: \texttt{\{\textcolor{red!80}{reasoning}\}}\\

        \bottomrule
    \end{tabularx}
    \label{app_tab:prompt_teacher}
\end{table}

\paragraph{Thinking Intervention Template}
\begin{table}[h]
    \centering
    \caption{Instruct template of streaming thinking intervention.}
    \label{Tab:Intervention_Temp}
    \renewcommand{\arraystretch}{1.2}
    \begin{tabularx}{\linewidth}{X}
        \toprule
        \rowcolor{gray!15} \multicolumn{1}{c}{\textit{Streaming thinking with direct answer}} \\
        \texttt{<|im\_start|>user\textbackslash n} {\{\textcolor{red!40}{Instruction \& Entire Input}\}}\texttt{<|im\_end|>\textbackslash n}\\
        \texttt{<|im\_start|>assistant\textbackslash n<think>}{\{\textcolor{red!40}{Streaming Thinking Content}\}}. \textcolor{blue!80}{Now, let me direct output the answer.}\\

        \midrule
        \rowcolor{gray!15} \multicolumn{1}{c}{\textit{Streaming thinking with global thinking}} \\
        \texttt{<|im\_start|>user\textbackslash n} {\textcolor{red!40}{\{Instruction \& Entire Input\}}}\texttt{<|im\_end|>\textbackslash n}\\
        \texttt{<|im\_start|>assistant\textbackslash n<think>}{\{\textcolor{red!40}{Streaming Thinking Content}\}}. \textcolor{blue!80}{Now, let me start the global thinking, focus on high-level reasoning trace that leads to the final answer.}\\

        \midrule
        \rowcolor{gray!15} \multicolumn{1}{c}{\textit{Streaming thinking with global thinking \& reflection}} \\
        \texttt{<|im\_start|>user\textbackslash n} {\{\textcolor{red!40}{Instruction \& Entire Input}\}}\texttt{<|im\_end|>\textbackslash n}\\
        \texttt{<|im\_start|>assistant\textbackslash n<think>}{\{\textcolor{red!40}{Streaming Thinking Content}\}}\textcolor{blue!80}{Now, let me start the global thinking.} {\{\textcolor{red!40}{Global Thinking Content}\}}. \textcolor{blue!80}{Wait, let me check again.}\\
        \bottomrule
    \end{tabularx}
    \label{app_tab:prompt_invention}
\end{table}

These prompts demonstrated in~\autoref{app_tab:prompt_invention} modulate how the assistant transitions from streaming thoughts (\texttt{<think>}) to an answer: (i) direct answer ends the streaming trace and outputs the result immediately; (ii) global thinking triggers a brief high-level consolidation before answering; (iii) global thinking \& reflection adds a final self-check pass. All variants preserve the streaming trace while explicitly steering when and how the final answer is produced.

\subsection{Quality Evaluation of Streaming CoT}
As described in the main text, we evaluate the quality of data generation using two metrics: \textit{granularity score} and \textit{sequential consistency score}. The granularity score ensures that each input sentence corresponds to a distinct and non-overlapping reasoning segment, while the sequential consistency score verifies that the reasoning process unfolds in the same order as the input sequence. In this appendix, we provide a detailed explanation of the two evaluation metrics.

The \textit{granularity score} measures the fine-grained alignment between the segmentation of the input context and that of the reasoning process. Formally, it is defined as the ratio between the number of boundary tokens
in the input and those in the output:
\[
\text{granularity} = \frac{N_{\text{EOS}}}{N_{\text{EOT}}},
\]
where $N_{\text{EOS}}$ and $N_{\text{EOT}}$ denote the counts of boundary tokens in the input and output, respectively. A score of 1 indicates that the reasoning preserves the segmentation of the input exactly, suggesting ideal boundary alignment.

The \textit{sequential consistency score} evaluates whether reasoning follows the input order in a streaming, segment-by-segment manner. We compute the semantic similarity between each input sentence and its corresponding reasoning segment:
\[
\text{consistency} = \text{sim}(R_t, C_t) = \frac{v_R \cdot v_C}{\|v_R\| \, \|v_C\|},
\]
where $v_R$ and $v_C$ denote the embedding vectors of reasoning sentence $R_t$ and input sentence $C_t$, respectively. We employ SentenceBERT~\citep{reimers2019sentence} to obtain embeddings. A higher score reflects that the reasoning faithfully preserves both the order and semantic content of the input stream. In cases where multiple reasoning sentences correspond to one input sentence, we
aggregate them as a single segment.

\subsection{Similarity Map of Streaming CoT}
\label{Appendix_Similarity}

\begin{figure}[h]
\begin{center}
\includegraphics[width=\linewidth]{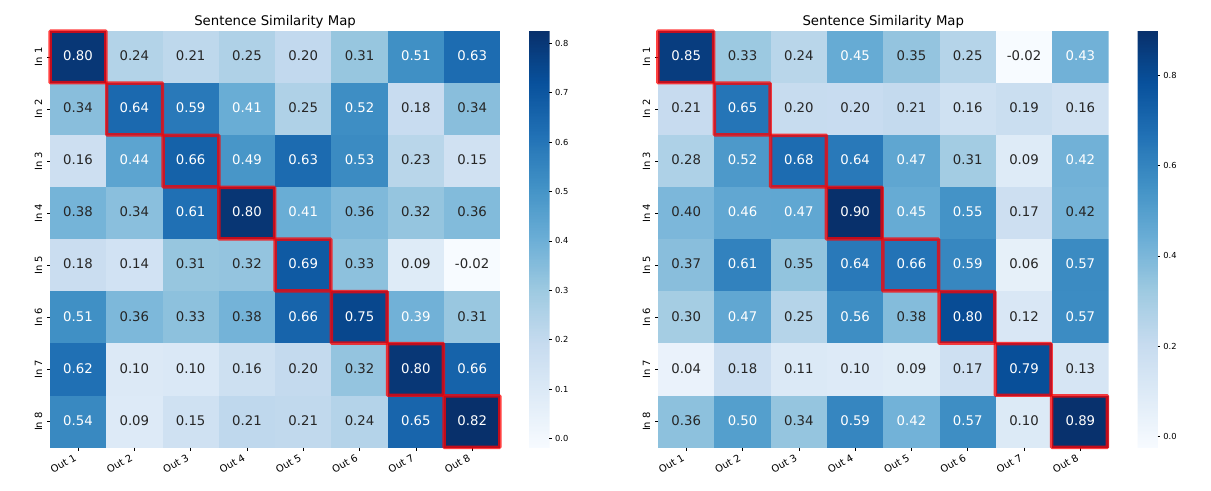}
\end{center}
\caption{Examples of sentence-level similarity map between the input sentences and the reasoning content (from math reasoning task). High similarity along the diagonal demonstrates strong alignment, enabling evaluation through sequential consistency.}
\label{Fig:sim_map}
\end{figure}

For further qualitative evidence, we provide a sentence-level similarity map in~\autoref{Fig:sim_map}, which visualizes how reasoning aligns with the input across time. This map highlights diagonal patterns when the reasoning process maintains strong sequential consistency.

\section{Dataset Details}
\label{Appendix_Dataset}
\paragraph{Math Reasoning}

We construct math reasoning data from GSM-Symbolic~\citep{mirzadehgsm}, MetaMathQA~\citep{yumetamath}, and TuLu-personas-math-grade~\citep{lambert2024tulu}.

GSM-Symbolic is a symbolic variant of GSM8K that reformulates arithmetic word problems into semi-structured symbolic expressions, explicitly preserving quantities, operators, and logical relations. It consists of two subsets, GSM-Symbolic-P1 (5K samples) and GSM-Symbolic-P2 (2.5K samples), where P2 includes longer and more complex problems. We randomly sample 4K and 2K instances from P1 and P2 for training, respectively, while the remaining data are used for evaluation.

MetaMathQA is an augmented dataset derived from the GSM8K and MATH training sets, containing 40K samples. It integrates symbolic reasoning with natural language explanation, covering multi-step arithmetic and algebraic reasoning tasks. We randomly select 1K samples for testing and use the remaining data for training instance generation.

TuLu-personas-math-grade comprises 50K math-related examples generated by GPT-4o under diverse instructional and persona conditions. The dataset emphasizes reasoning diversity and linguistic variation, providing rich supervision signals for reasoning style adaptation. All samples are used for generating training data.

To align with the streaming objective of “thinking while reading,” we retain inputs with at least three sentences and exclude tasks with very short inputs (e.g., proofs), which emphasize deep reasoning rather than streaming reasoning.
We mix the generated data from all sources, and after filtering for correctness and evaluating the quality of streaming CoT, we retain 7.9K samples for training.

\paragraph{Logical Reasoning}
For logical reasoning, we evaluate on ProofWriter~\citep{tafjord2020proofwriter} and LogicNLI~\citep{tian2021diagnosing}.
ProofWriter is a benchmark designed for evaluating multi-step logical reasoning and theorem proving capabilities in natural language. It is derived from the EntailmentBank corpus, where each example consists of a hypothesis and a set of supporting facts expressed in natural language. The task requires the model to infer whether the hypothesis is entailed, contradicted, or neutral given the supporting facts, while optionally generating an explicit reasoning chain that connects the premises to the conclusion.
LogicNLI is a natural language inference (NLI) benchmark designed to evaluate the logical reasoning ability of language models beyond surface-level semantics. Each example in LogicNLI consists of a premise and a hypothesis pair, annotated with one of three logical relations: entailment, contradiction, or neutral. Unlike conventional NLI datasets that focus primarily on lexical or syntactic cues, LogicNLI emphasizes reasoning over formal logic structures such as conjunction, disjunction, negation, quantifiers, and implication.

The ProofWriter dataset contains approximately 40K samples. We randomly sample 1K instances for testing, while the remaining data are used for generating training instances.
For LogicNLI, we use its original 1k test set, while the remaining 16K training data is used for generating streaming CoT data. After applying the streaming CoT generation process, the resulting dataset contains approximately 20k instances for training.

\paragraph{Context-based QA Reasoning}
For context-based QA reasoning task, we evaluate StreamingThinker on PubMedQA~\citep{jin2019pubmedqa} and HotpotQA~\citep{yang2018hotpotqa} datasets.

PubMedQA is a biomedical question answering benchmark designed to evaluate factual reasoning and evidence-based inference in scientific texts. Each instance consists of a biomedical research question, a context paragraph extracted from PubMed abstracts, and a short-form answer annotated as yes, no, or maybe. The dataset emphasizes reasoning over factual statements, experimental findings, and logical connections.

HotpotQA is a large-scale question answering dataset designed to evaluate multi-hop reasoning across multiple supporting documents. Each example consists of a question, a set of supporting paragraphs from Wikipedia, and a corresponding answer. Unlike single-hop QA datasets that require reasoning within a single sentence or passage, HotpotQA explicitly requires models to integrate information from multiple contexts to derive the correct answer.

The {PubMedQA} dataset contains 1K samples for testing and 61K samples for training, where all of training samples are used for streaming CoT generation.
The HotpotQA dataset includes 90K samples. We randomly sample 1K examples for testing and 60K for streaming CoT generation.
After correctness filtering and streaming CoT quality evaluation, we retain a total of 16K samples for training.

\section{Model Details}
\subsection{Training Details}

During training, we adopt a streaming mask region on top of the standard decoder-only LLM causal mask to enforce strict streaming constraints. Specifically, each training instance is composed of grouped \emph{input} and \emph{reasoning} segments, where the streaming mask (Eq.~\ref{eq:mask}) ensures that reasoning at time $t$ cannot attend to future inputs. Additionally, \textit{the position encoding} of both the input and CoT is modified to group-wise streaming encoding, preventing positional conflicts.

For training, we set the batch size to 16 and employ the AdamW optimizer with a learning rate of 1e-5. We additionally enable activation checkpointing to mitigate memory overhead from maintaining parallel caches during training. These choices align the training setup with prior LLM pretraining while introducing only the modifications necessary for streaming compatibility.

\subsection{Decoding Strategy}
During inference, decoding must also respect the streaming paradigm. Unlike standard autoregressive generation where all inputs are visible, our decoding operates in a \emph{streaming-aware} manner. Input tokens are continuously appended to a dedicated input cache, while reasoning tokens are generated in parallel using a separate reasoning cache. At each step, the streaming mask ensures that reasoning about the current sentence depends only on past inputs and past reasoning, never on unseen future inputs.

For generation, we follow the sampling parameters of Qwen3. This balances fluency and diversity while maintaining consistency across sequentially produced reasoning sentences. Importantly, because input and reasoning caches are maintained independently, decoding proceeds with lower latency: new inputs can be consumed without flushing or rewriting the reasoning cache, and reasoning updates can be emitted as soon as the corresponding input sentence completes. This strategy preserves alignment with streaming training while ensuring responsiveness in real-time scenarios. The streaming thinking decoding process is shown in Algorithm \autoref{alg:decoding}.

\begin{algorithm}[t]

\caption{Streaming thinking with parallel KV caches}
\textbf{Input:} \parbox[t]{0.8\linewidth}{Source length list $S$, target length list $T$.}
\begin{spacing}{1.2}
\begin{algorithmic}[1]
\State \parbox[t]{\linewidth}{Initialize input KV cache $I_{cache}$, output KV cache $O_{cache}$, and merged KV cache $M_{cache}$.}
\State Read the first input sentence, and save hidden state to input KV cache $I_{cache}$.
\State \texttt{[Parallel] For Input KV Cache (prefill):}
\While{Input sentence is arrival}:
    \State Separate $M_{cache}$ to $I_{cache}$ and $O_{cache}$.
    \State Read the input sentence, and save hidden state to input KV cache $I_{cache}$.
\EndWhile
\State \texttt{[Parallel] For Output KV Cache (decoding):}
\While{Input sentence is arrival}:
    \State \textbf{if} $N_{EOS}\geq N_{EOT}+1$: \textcolor{red!50}{Sentence arrival time is less than LLM decoding time.}
    \State \qquad Select a slice of the input KV cache $I'_{cache}$ to keep $N_{EOS}==N_{EOT}+1$.
    \State \textbf{elif} $N_{EOS}< N_{EOT}+1$:  \textcolor{red!50}{Sentence arrival time  exceeds LLM decoding time.}
    \State \qquad Wait for the input KV cache prefilling and keep $N_{EOS}==N_{EOT}+1$.
    \State \qquad Select a slice of the input KV cache $I'_{cache}$.
    \State  Merge $I'_{cache}$ and $O_{cache}$ to $M_{cache}$.
    \State  Decode the streaming thinking tokens and save to $M_{cache}$.
\EndWhile
\State \texttt{[End of Parallel]}
\State Generate controllable deep thinking with thinking intervention.
\State \textbf{Return:} The streaming CoT.

\end{algorithmic}
\end{spacing}
\label{alg:decoding}
\end{algorithm}

\subsection{Relation between Parallel KV Caches and Prefill-Decode Separation}
While our parallel KV caches mechanism may formally resemble techniques for Prefill-Decode separation~\citep{zhong2024distserve,qin2024mooncake}, the foundational motivations and operational goals behind the two approaches are fundamentally distinct.

Prefill-Decode separation is primarily a system-level optimization designed to enhance throughput in batched inference. It bifurcates the generation process into two discrete phases: a computationally intensive prefill stage that processes the input prompt in a parallel batch, and a memory-bandwidth-bound decoding stage for auto-regressive token generation. The core objective is to maximize hardware utilization by applying specialized computational kernels to each distinct phase, thereby {improving efficiency for a fundamentally sequential task}.

In contrast, the design of our parallel KV caches is driven by the goal of enabling the model to generate output concurrently while still consuming input. In other words, our approach is designed {to overlap the processing of incoming data with the generation of the output}, breaking the strict sequential dependency where generation can only begin after the entire input has been processed.

\section{Evaluation Metric}
\label{Appdenx_Metric}

\subsection{Reasoning Accuracy}
\paragraph{Pass@1 Accuracy} This metric measures the proportion of test instances in which the model’s top-1 generated output exactly matches the ground-truth answer. Unlike relaxed metrics that consider multiple sampled generations (e.g., pass@$k$), pass@1 provides a strict estimate of the model’s reliability under single-shot decoding, which is the default usage setting for most real-world applications. A higher pass@1 score thus indicates stronger reasoning consistency and correctness without relying on sampling diversity.

\subsection{Reasoning Latency}

\paragraph{Token-length Delay}
In streaming tasks, token-length delay measures \emph{how many tokens must be consumed before the model begins its reasoning process}. Formally, it is defined as the gap between the start of the input stream and the position of the first reasoning token. A smaller token delay indicates that the model can initiate reasoning earlier in the stream, leading to faster overlap between input consumption and output generation. This metric thus reflects the responsiveness of reasoning onset in the streaming paradigm.
\begin{equation}
\text{Token-Delay} =
\text{Position of first reasoning token}
\end{equation}

\begin{figure}[h]
\begin{center}
\includegraphics[width=\linewidth]{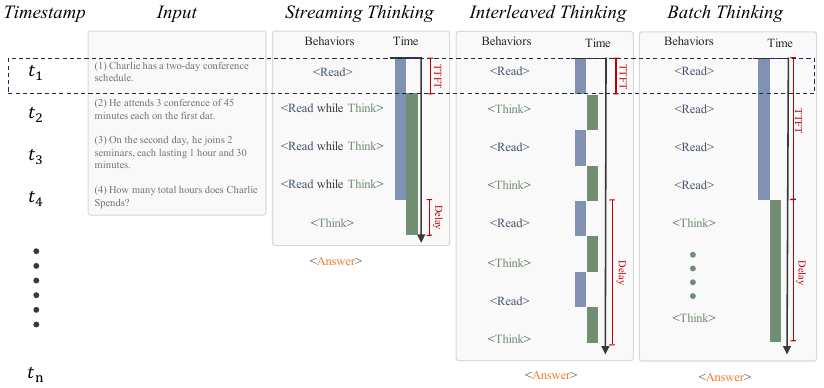}
\end{center}
\caption{Comparison of reasoning paradigms. Streaming thinking enables concurrent reading and reasoning with minimal delay, while interleaved thinking alternates between the two but still accumulates overhead, and batch thinking postpones reasoning until all inputs are read, leading to the largest delay.}
\label{Fig:app_delay}
\end{figure}

\paragraph{Time Delay}
Time delay measures the real-time latency \emph{between the arrival of the last input token and the emission of the first answer token}. This metric captures system-level responsiveness, incorporating factors such as decoding speed, cache management, and hardware throughput. While token delay evaluates the generation behavior in discrete units, time delay provides a wall-clock perspective that better reflects the user’s perceived waiting time. The \autoref{Fig:app_delay} illustrates the latency comparison between streaming thinking and batch thinking.

\end{document}